\documentclass[runningheads]{llncs}

\usepackage[mobile]{eccv}

\usepackage{eccvabbrv}

\usepackage{graphicx}
\usepackage{booktabs}

\usepackage[accsupp]{axessibility}  %
\usepackage{lipsum}

\usepackage{changepage}
\usepackage{multirow}
\usepackage{anyfontsize}
\usepackage{graphicx}
\usepackage{fancybox}

\usepackage[pagebackref,breaklinks,colorlinks]{hyperref}

\usepackage{orcidlink}

\usepackage[normalem]{ulem}

\begin{document}

\title{RoDUS: Robust Decomposition of Static and Dynamic Elements in Urban Scenes} 

\titlerunning{RoDUS}

\author{Thang-Anh-Quan Nguyen\orcidlink{0009-0004-4001-4279} \and
Luis Rold{\~a}o\orcidlink{0000-0003-0482-3584} \and 
Nathan Piasco\orcidlink{0000-0001-7952-6643} \and
\\Moussab Bennehar\orcidlink{0000-0002-6566-6132} \and 
Dzmitry Tsishkou\orcidlink{0009-0002-9798-3316}}

\authorrunning{T.A.Q.~Nguyen et al.}

\institute{Noah's Ark, Huawei Paris Research Center, France\\
\email{\{thang.anh.nguyen,luis.roldao,nathan.piasco,\\moussab.bennehar,dzmitry.tsishkou\}@huawei.com}}

\maketitle

\begin{figure}%
  \centering
  \includegraphics[width=\linewidth]{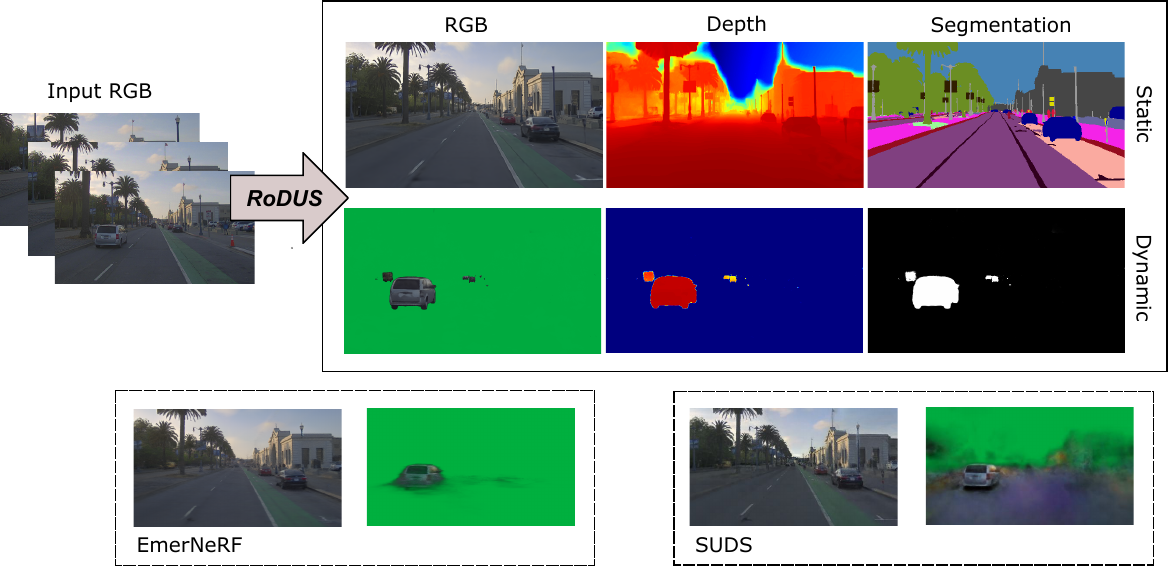}
  \caption{RoDUS is a neural scene representation designed to decompose 4D dynamic scenes into two elements: moving foreground and static background. This decomposition remains consistent across all photometric, geometric, and semantic aspects.}
  \label{fig:teaser}
\end{figure}

\vspace{-0.5cm}
\begin{abstract}
    The task of separating dynamic objects from static environments using NeRFs has been widely studied in recent years.
    However, capturing large-scale scenes still poses a challenge due to their complex geometric structures and unconstrained dynamics.
    Without the help of 3D motion cues, previous methods often require simplified setups with slow camera motion and only a few/single dynamic actors, leading to suboptimal solutions in most urban setups.
    To overcome such limitations, we present RoDUS, a pipeline for decomposing static and dynamic elements in urban scenes, with thoughtfully separated NeRF models for moving and non-moving components. Our approach utilizes a robust kernel-based initialization coupled with 4D semantic information to selectively guide the learning process.
    This strategy enables accurate capturing of the dynamics in the scene, resulting in reduced floating artifacts in the reconstructed background, all by using self-supervision.
    Notably, experimental evaluations on KITTI-360 and Pandaset datasets demonstrate the effectiveness of our method in decomposing challenging urban scenes into precise static and dynamic components.
    \keywords{Neural Radiance Fields (NeRFs) \and Static-Dynamic Decomposition \and Motion Segmentation}
\end{abstract}

\section{Introduction}
\label{sec:intro}

There have been numerous works that achieved success in various 3D representations including both implicit~\cite{mildenhall2021nerf} and explicit~\cite{schonberger2016structure, schonberger2016pixelwise} scene modeling.
Therefore, reconstructing and decomposing large-scale dynamic outdoor scenes is becoming a critical task for 3D scene understanding, especially for domains such as closed-loop simulation for autonomous driving~\cite{wu2023mars, yang2023unisim, tonderski2023neurad}.
Accurate segmentation and removal of dynamic objects are crucial as they are the main reason for artifacts during novel view synthesis (NVS) of background regions due to view inconsistency. 
This inevitably affects other downstream applications. For instance, several works~\cite {liu2023robust, deka2023erasing, herau2023soac} require masking out eventual dynamic objects as they estimate camera poses. Indeed, in highly dynamic scenes, without feeding the correct masks, they tend to make wrong refinements of the poses.
On the other hand, dynamic objects often contain valuable information in the scene and play a main role in tasks such as surveillance and video compression~\cite{mattheus2020review}.

However, advancements are lagging as understanding time-dependent scenes is an ill-posed problem. This is because it demands considerable efforts to detect and treat moving components separately while ensuring spatiotemporal consistency compared to static scenes~\cite{park2023temporal, wang2023masked}.
The challenges further arise in driving settings, featuring not only large unbounded scenes, with lots of unconstrained dynamics and various illuminations but also large temporal gaps captured from the ego-vehicle. This demonstrates that a majority of the background regions are captured only for a short duration, due to high occlusions caused by other dynamic objects while the ego-vehicle is also in continuous motion.
Unlike previous works~\cite{park2021nerfies, park2021hypernerf, li2021neural, wu2022d}, where scenes typically involve a single object with slow ego-motion, the scenarios encountered in driving sequences are less controllable.

Until now, the research community has proposed various solutions to the challenge of decoupling dynamic objects from static environments. Many develop NeRF-based multi-branch architectures in which only the dynamic (\ie time-dependent) model is employed for rendering dynamic objects while the background is learned using a separate static model.
Subsequently, these methods need to detect moving objects offline, so that the static model could later avoid sampling points at regions that belong to dynamic objects~\cite{ost2021neural, kundu2022panoptic, wu2023mars, xu2023discoscene, yang2023urbangiraffe, zhou2023drivinggaussian, yan2024street}. Nevertheless, they end up relying on expensive annotations for segmenting and tracking every object in the scene to enable correct guidance. Such limitation restricts their adaptability to many datasets. Others~\cite{martin2021nerf, chen2022hallucinated} learn to classify the inconsistency through a rejection factor while jointly optimizing both branches.
However, NeRF's view-dependent capacity makes it difficult to distinguish non-Lambertian but static from true dynamic objects, requiring careful and expensive parameter tuning to reach optimal separation.

In this paper, by building upon self-supervised 4D scene representation using NeRF, RoDUS (\textbf{Ro}bust \textbf{D}ecomposition of Static and Dynamic Elements in \textbf{U}rban \textbf{S}cenes) employs a two-pathway architecture with additional sky, road, and shadow modeling that is suitable for urban scenes. While the architecture has been mainly used for representing dynamic scenes as a whole, we argue that it appears a possibility of both branches learning incorrect information (\ie local minima), as long as the overall reconstruction satisfies the ground truth image. This deviates from the initial design intent, where the static model is meant to learn only the static background, and vice versa. Toward this goal, we draw inspiration from~\cite{barron2019general} by proposing a robust kernel-based strategy, thus reducing L2 sensitivity on the static model. Additionally, we introduce a view-independent semantic field to prevent the dynamic model from expressing unwanted regions, while still being able to refine high-frequency detail. This semantic awareness is further used to guide the information flow, thus balancing the contribution of each branch. 
To summarize, the main contributions of our work are:
\begin{itemize}
    \item We demonstrate RoDUS's capability of static-dynamic decomposition across photometric, geometric, and semantic outputs in a self-supervised manner.
    \item We introduce a robust initialization scheme and semantic reasoning ability with a masking mechanism that ensures the quality of separation.
    \item We conduct comprehensive comparisons on several challenging driving sequences, indicating our RoDUS's performance on static-dynamic components extraction, and decoupling compared to other SoTA methods.
\end{itemize}

\section{Related Work}
Recently, many works have achieved promising results in high-fidelity view synthesis~\cite{barron2022mip} and combining with semantic~\cite{fu2022panoptic, kundu2022panoptic, zhang2023nerflets, irshad2023neo}, foundation model-guided~\cite{kobayashi2022decomposing, tschernezki2022neural} reasoning to better accommodate urban driving scenarios~\cite{rematas2022urban, guo2023streetsurf}. As most of the above methods focus on static scenes, some other works have targeted non-static, transient elements from video sequences~\cite{park2021nerfies, park2021hypernerf, li2021neural, li2022neural, park2023temporal, wang2023masked}.

\noindent\textbf{NeRFs for outliers rejection.}
NeRF-W~\cite{martin2021nerf} is the pioneering work that decomposes the scene into static and transient components to reconstruct landmarks from unconstrained phototourism datasets. They propose per-frame latent embeddings and a transient head to model different lighting conditions and transient effects, thus these phenomena can be removed from the rendering function at inference.
Alternatively, RobustNeRF~\cite{sabour2023robustnerf} successfully rejects distractors from the scenes using a trimmed kernel to modify the photometric loss into an Iteratively Least Squares problem, they also include patch filtering to maintain sharp reconstruction quality. 
Ha-NeRF~\cite{chen2022hallucinated}, instead, parameterizes the kernel as learnable and uses 2D visibility map per image as a weight function. This occlusion-aware module is proven to be more accurate at separating transients.

\noindent\textbf{Motion segmentation.}
In addition to advancements in 3D vision, there has been extensive research focusing on self-supervised motion segmentation at 2D image level. The majority of these methods rely on 2D trackers~\cite{deka2023erasing}, optical flow patterns~\cite{cao2019learning, li2021unsupervised, ye2022deflowslam} and/or image warping~\cite{saunders2023dyna} to distinguish between ego motions and object motions and segment them.
These approaches have clear limitations as they focus solely on image-level processing and lack scene-level understanding. Therefore, they struggle to handle complicated scenes with large camera motion~\cite{wu2022d}.
Furthermore, those who follow the pipeline of \textit{first-process-then-NeRF} create not only label ambiguity that affects learning but also add extra costs for separate perception processing tasks, considering that motion segmentation or flow classification themselves are already challenging. 

\noindent\textbf{NeRFs for dynamic scene reconstruction.}
To represent dynamic scenes, a direct solution is to add time as an additional input~\cite{pumarola2021d, li2022neural} or learn a deformation field that maps the input coordinates into a canonical space~\cite{park2021nerfies, park2021hypernerf}. These time-dependent extensions, however, fail to accurately represent complex scenes and are unable to disentangle the dynamic components. Several works have adopted the idea in multi-branch architectures (\eg NSFF~\cite{li2021neural}, NeuralDiff~\cite{tschernezki2021neuraldiff}, STaR~\cite{yuan2021star}, D$^2$NeRF~\cite{wu2022d}), decomposing the scene using different radiance fields.
Nonetheless, the mentioned methods lack robustness and are not tested in urban driving scenarios where it is much harder to manage the number of dynamic objects within the scene. Consequently, their performance significantly degrades due to the blurriness of dynamic objects and floating artifacts in background regions, restricting their applicability.
Follow-up works exploit object-centric approaches, especially NGS~\cite{ost2021neural}, PNF~\cite{kundu2022panoptic}, and DisCoScene~\cite{xu2023discoscene} express dynamic scenes and target automotive data, but they are fundamentally constrained by per-scene 3D annotations including bounding boxes and instance tracking which are difficult to obtain reliably without 3D sensors and proper calibration~\cite{herau2023moisst}. 

SUDS~\cite{turki2023suds} and EmerNeRF~\cite{yang2023emernerf} are the most similar to our approach since they utilize 2D off-the-shelf models to enable object-/scene-level semantic understanding while still being able to separate dynamic objects. 
However, our approach differs from existing works in that we place greater emphasis on the representation of each branch individually, a factor often overlooked by other studies. This focus proves advantageous in understanding complex scenes. Specifically, our dynamic model is designed to learn only the true dynamic instances rather than capturing variations in illumination or pixel changes caused by ego-motion.

\section{Method}
\begin{figure}[tb]
  \centering
  \includegraphics[width=\linewidth]{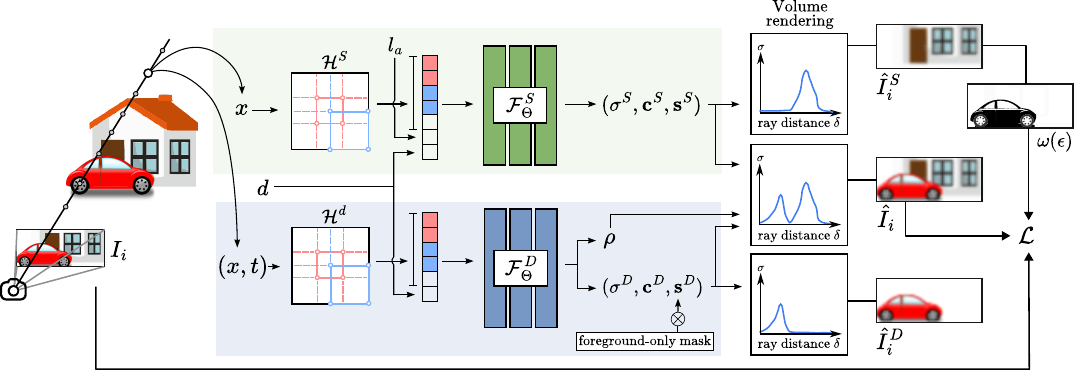}
  \caption{\textbf{RoDUS's architecture}.
  Our model comprises two separate branches that take as input sampled positions $\mathbf{x}$, viewing direction $\mathbf{d}$, and their timestamp $t$, and generate outputs for every query coordinate.
  Each branch is represented by a separate hash grid and the MLP-based neural function, which predicts colors, densities, and semantics.
  The rendered static RGB is used to calculate IRLS map $\omega(\epsilon)$ during the robust initialization step (\cref{sec:loss}), while dynamic semantic outputs are passed through a ``foreground-only mask'' to prevent over-explaining background regions (\cref{sec:semantic}).}
  \label{fig:architecture}
\end{figure}

This section details RoDUS's methodology of disentangling and composing representation with its computational pipeline illustrated in \cref{fig:architecture}. 
\cref{sec:representation} describes the architecture for our scene representation.
\cref{sec:semantic} introduces the semantic-radiance field used in RoDUS, how it can be rendered using differentiable volume rendering and contributes to the separation.
Finally, we formulate the losses to train RoDUS in \cref{sec:loss} as well as our proposed training strategy in \cref{sec:strategy}.

\noindent\textbf{Goals.} Our main goal is to learn a scene representation of a dynamic environment (\eg a driving scene) with the ability to semantically disentangle between static background (\eg road, building, vegetation, parked vehicles) and dynamic objects in a self-supervised manner.
RoDUS takes a set of $N$ input RGB images, their associated poses, and capturing timestamps $\{I_i, p_i, t_i\}_{i=1}^{N}$.
Additionally, our model relies on semantic $\{S_i\}_{i=1}^{N}$, and geometric cues obtained from pre-trained 2D segmentation model and LiDAR depth measurements.
At inference, we wish to achieve rendered images for static $\{\hat{I}_i^S, \hat{S}_i^S\}_{i=1}^{N}$ and dynamic $\{\hat{I}_i^D, \hat{S}_i^D\}_{i=1}^{N}$ components of the scene for each timestamp independently, which can also be combined for a complete view. 
As mentioned above, we study the quality of individual components rather than conforming solely to a satisfying composed result.

\subsection{Scene Representation}
\label{sec:representation}

\textbf{Preliminaries.} RoDUS is based on NeRF~\cite{mildenhall2021nerf}, which is designed to capture both geometry and view-dependent appearance of a scene from a set of RGB images. It encodes the scene within a Multi-Layer Perceptron (MLP) and outputs color $\mathbf{c}(\mathbf{x}, \mathbf{d}) \in \mathbb{R}^3$ and density $\sigma(\mathbf{x}) \in \mathbb{R}$ at any query position $\mathbf{x} \in \mathbb{R}^3$ along with its viewing direction $\mathbf{d} \in \mathbb{R}^3$. During training, using intrinsic parameters and poses, every pixel of the image undergoes a ray marching procedure $\mathbf{r}$
to sample inputs for the MLP. The model parameters are optimized to predict the colors rendered along the ray using the mean squared error: $\mathcal{L} = \|\hat{\mathbf{C}}_i(\mathbf{r}) - \mathbf{C}_i(\mathbf{r})\|_2^2$.

\noindent\textbf{Architecture}.
Inspired by~\cite{martin2021nerf, li2021neural, wu2022d}, we factorize the scene into two elements, each learned through a separate pathway. Each branch is modeled independently using its own multi-level hash grid representation and MLPs. This approach enables the concurrent learning of both aspects of the scene.

The static branch uses a 3D hash grid $\mathcal{H}^S$ for positions and positional embedding $\gamma(\cdot)$ for viewing directions, analogous to the one used in~\cite{muller2022instant}.
To model lighting variations, we also condition  on a per-frame latent embedding $l_a$~\cite{martin2021nerf}:
\begin{equation}
    (\sigma^S, \mathbf{c}^S, \mathbf{s}^S)=\mathcal{F}^S_\Theta(\mathcal{H}^S(\mathbf{x}), \gamma(\mathbf{d}), l_a).
\end{equation}

As the dynamic branch allows its outputs to depend on time, we extend to 4D hash grid $\mathcal{H}^D$ with one additional normalized timestamp $t \in [0, 1]$ as input. 
We additionally incorporates a shadow head~\cite{wu2022d}, designed to learn a shadow ratio $\rho\in [0, 1]$. This scalar is later used to scale down the static radiance values $\mathbf{c}^S$ in \cref{eq:color_render}:
\begin{equation}
\begin{split}
    (\sigma^D, \mathbf{c}^D, \mathbf{s}^D) &= \mathcal{F}^D_\Theta(\mathcal{H}^D(\mathbf{x}, t), \gamma(\mathbf{d})),\\
    \rho &= \mathcal{F}^\rho_\Theta(\mathcal{H}^D(\mathbf{x}, t)).
\end{split}
\end{equation}

\noindent\textbf{Rendering.}  
We calculate the color $\hat{C}(\mathbf{r})$ of each ray $\mathbf{r}$ by integrating all the sampled points along the ray:
\begin{equation}
    \hat{C}(\mathbf{r})=\sum_{i=1}^{K}T_i\left(\alpha_i^S (1-\rho_i)\mathbf{c}_i^S + \alpha_i^D \mathbf{c}_i^D\right),
\label{eq:color_render}
\end{equation}

\noindent where $T_i=\exp\left(-\sum_{j=1}^{i-1}(\sigma_j^S+\sigma_j^D)\delta_j\right)$ is the accumulated transmittance and $\alpha_i^\xi=1-\exp(-\sigma_i^\xi\delta_i)$ with $\xi\in\{S, D\}$ are alpha values. Given the outputs of each branch, we can also render static and dynamic maps separately.

\subsection{Enabling Semantic Awareness}
\label{sec:semantic}
NeRF provides low-level implicit representations of geometry and radiance but lacks a higher-level understanding of the scene. It is shown in~\cite{sabour2023robustnerf} that relying only on view-dependent radiance and sensitive mean square error makes the model prone to outliers and over-fit observations. 
On the other hand, semantic labels can be conceptualized as a \textit{view-invariant} function~\cite{zhi2021place}, mapping world coordinates $\mathbf{x}$ to a distribution over semantic labels $\mathbf{s}(\mathbf{x}) \in \mathbb{R}^L$, where $L$ is the number of classes. We choose to integrate a semantic head to predict the semantic class for each point, similar to the color head. Subsequently, the same rendering formula is applied for semantics to achieve 2D semantic segmentation maps:
\begin{equation}
    \hat{S}(\mathbf{r}) = \sum_{i=1}^{K}T_i\left(\alpha_i^S\mathbf{s}_i^S + \alpha_i^D \mathbf{s}_i^D\right).
\end{equation}

The use of semantic supervision in RoDUS serves two main purposes. First, it provides assumptions for multiple classes, forming the basis for the loss formulations (refer to \cref{sec:loss}) that contribute to enhancing rendering quality. Second, it enables the model to capture detailed scene layout and object boundary information with a high level of precision thereby assisting in decomposition.

\begin{figure}[tb]
  \centering
    \begin{subfigure}[b]{0.64\linewidth}
        \includegraphics[width=\linewidth]{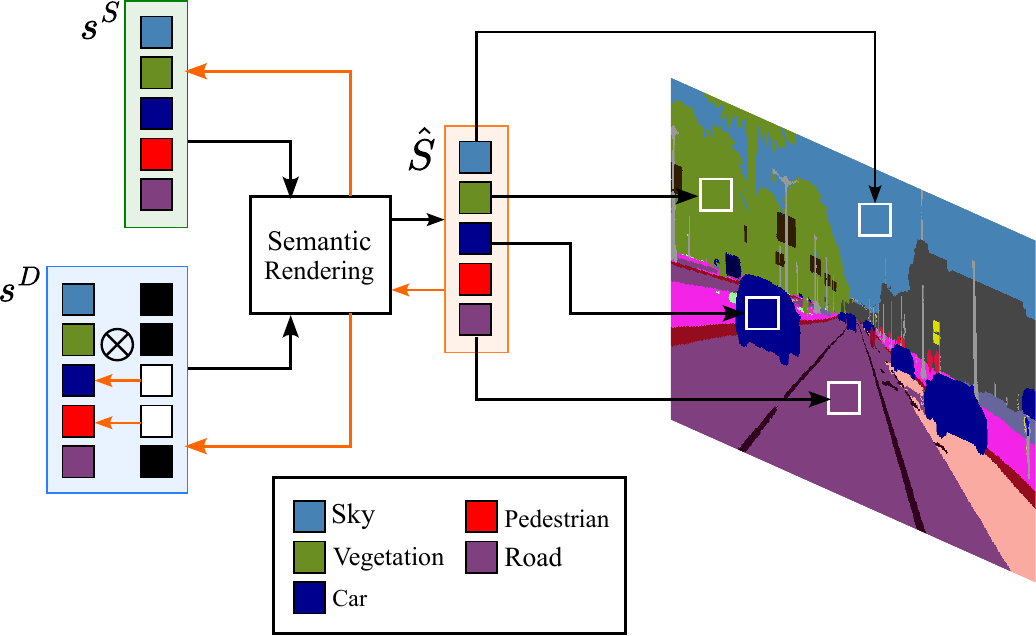}
        \caption{}
    \end{subfigure}
    \begin{subfigure}[b]{0.34\linewidth}
    \includegraphics[width=0.95\linewidth]{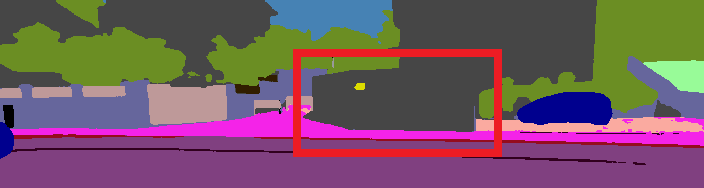}
    \includegraphics[width=0.95\linewidth]
    {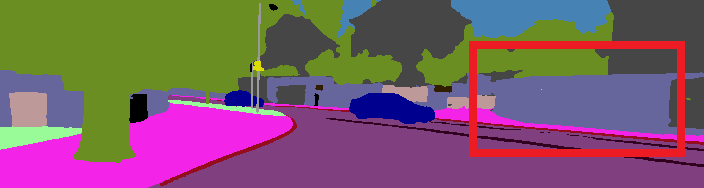}
    \caption{}
    \includegraphics[width=0.95\linewidth]{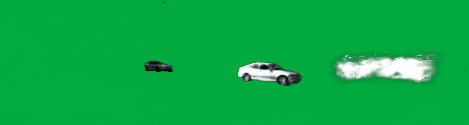}
    \caption{}
    \includegraphics[width=0.95\linewidth]{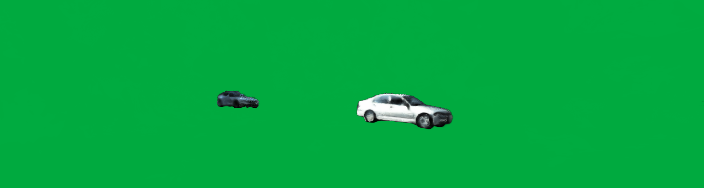}
    \caption{}
    \end{subfigure}

  \caption{(a) We enforce a ``foreground-only mask'' to the dynamic semantic head. In the forward pass (black arrows), the mask prevents the dynamic branch from outputting pixels that do not belong to foreground classes. While in the backward pass (\textcolor{orange}{orange arrows}), it restricts the dynamic field from learning background pixels, which may primarily come from noisy annotations.
  (b) Predictions generated by 2D segmentation model are noisy and inconsistent between views. (c) Since the dynamic field includes a temporal dimension, it ends up learning these conflicting labels, satisfying the overall loss. (d) Therefore, our proposed mask is used to tackle the problem.}
  \label{fig:semantic_gradient_flows}
\end{figure}

One point to consider is that supervision for segmentation with noisy pseudo labels should not impact the underlying geometry. Otherwise, the model can still satisfy conflicting labels by changing the dynamic weights, resulting in the wrong geometry. 
Previous works~\cite{fu2022panoptic, siddiqui2023panoptic} stop gradients from the semantic head back to the densities to solve this problem. However, we believe that semantic information plays an important role in correcting the separation. We instead limit the dynamic model from contributing static class prediction to the final results by putting a mask in the last layer of its semantic head. This masking strategy can be viewed as a form of gradient stopping the learning but in a selective way as illustrated in~\cref{fig:semantic_gradient_flows}. 
Note that we do not have motion information for each instance, we only leverage the fact that certain classes are potentially movable to penalize the dynamic model from predicting background (non-movable) samples.

We apply \texttt{softmax} normalization at each query point individually before ray integration, as introduced in \cite{siddiqui2023panoptic}. This is done due to the unbounded nature of logits, ensuring that the semantic head can not generate excessively high scores in regions with low density to overcome high-density areas. Remarkably, this not only improves the stability of rendering semantics but also prevents half-precision floating point (\texttt{float16}) overflowing when working with frameworks such as \texttt{tiny-cuda-nn}~\cite{muller2021tinycudann}. By performing inferences in 3D space and rendering to 2D with a foreground masking mechanism, our semantic outputs are multi-view consistent by design.

\subsection{Loss Formulation}
\label{sec:loss}
\textbf{Reconstruction losses.} 
Consistent with other NeRF works, we incorporate both L2 and DSSIM~\cite{wang2004image} losses to supervise the pixel color $\mathcal{L}_{rgb}$. In cases where LiDAR data is available, depth supervision~\cite{deng2022depth, rematas2022urban} loss $\mathcal{L}_{d} = \|\hat{d}_i(\mathbf{r}) - d_i(\mathbf{r})\|_2^2$ is considered to enhance the reconstruction accuracy.
In addition, cross-entropy~\cite{zhi2021place} loss is applied to provide supervision on the generated semantic labels derived from pseudo-ground truth segmentation, expressed as $\mathcal{L}_{sem} = CE(\hat{S}(\mathbf{r}), S(\mathbf{r}))$.

\noindent\textbf{Robust loss.}
Our objective is to reduce the sensitivity of the static model on dynamic components through a robust photometric loss. This loss can be formulated as an Iteratively Reweighted Least Squares (IRLS) problem:
\begin{equation}
    \mathcal{L}_{robust} = \kappa\left(\|\hat{C}^S(\mathbf{r}) - C(\mathbf{r})\|_2\right) = \omega(\epsilon(\mathbf{r}))\cdot\|\hat{C}^S(\mathbf{r}) - C(\mathbf{r})\|_2^2,
    \label{eq:robust_loss}
\end{equation}

\noindent where $\epsilon(\mathbf{r})=\|\hat{C}^S(\mathbf{r}) - C(\mathbf{r})\|_2$ is the error residuals and $\omega(\epsilon)=\epsilon^{-1}\cdot\partial\kappa(\epsilon)/\partial\epsilon$ is the weight function of the residuals at the previous iteration. 
Among the family of kernels~\cite{barron2019general}, we follow Sabour \etal~\cite{sabour2023robustnerf} and choose a trimmed estimator~\cite{chetverikov2002trimmed} which classifies outliers based on a percentile threshold $\mathcal{T}_\epsilon$ calculated from the whole batch of rays. This choice ensures outliers are completely removed (\ie hard weights). Furthermore, as every object has neighbor connections, we apply a spatial filter $\mathcal{F}$ on the residuals map to capture the smoothness, ensuring local support of the kernel:
\begin{equation}
    \omega(\epsilon(\mathbf{r})) = \mathcal{F}\left[\epsilon(\mathbf{r})\leq\mathcal{T}_\epsilon\right],
\end{equation}

\noindent where $\mathcal{F}$ consists of a box filter to diffuse inlier/outlier labels and a sub-patch level consistency filter that takes into account the rejection behavior of its surrounding neighborhoods. This filter aims to remove high-frequency details from being classified as outliers and allows them to be captured by the NeRF models.

\begin{figure}[tb]
    \centering
    \begin{subfigure}{0.19\textwidth}
    \centering
        \includegraphics[width=\linewidth]{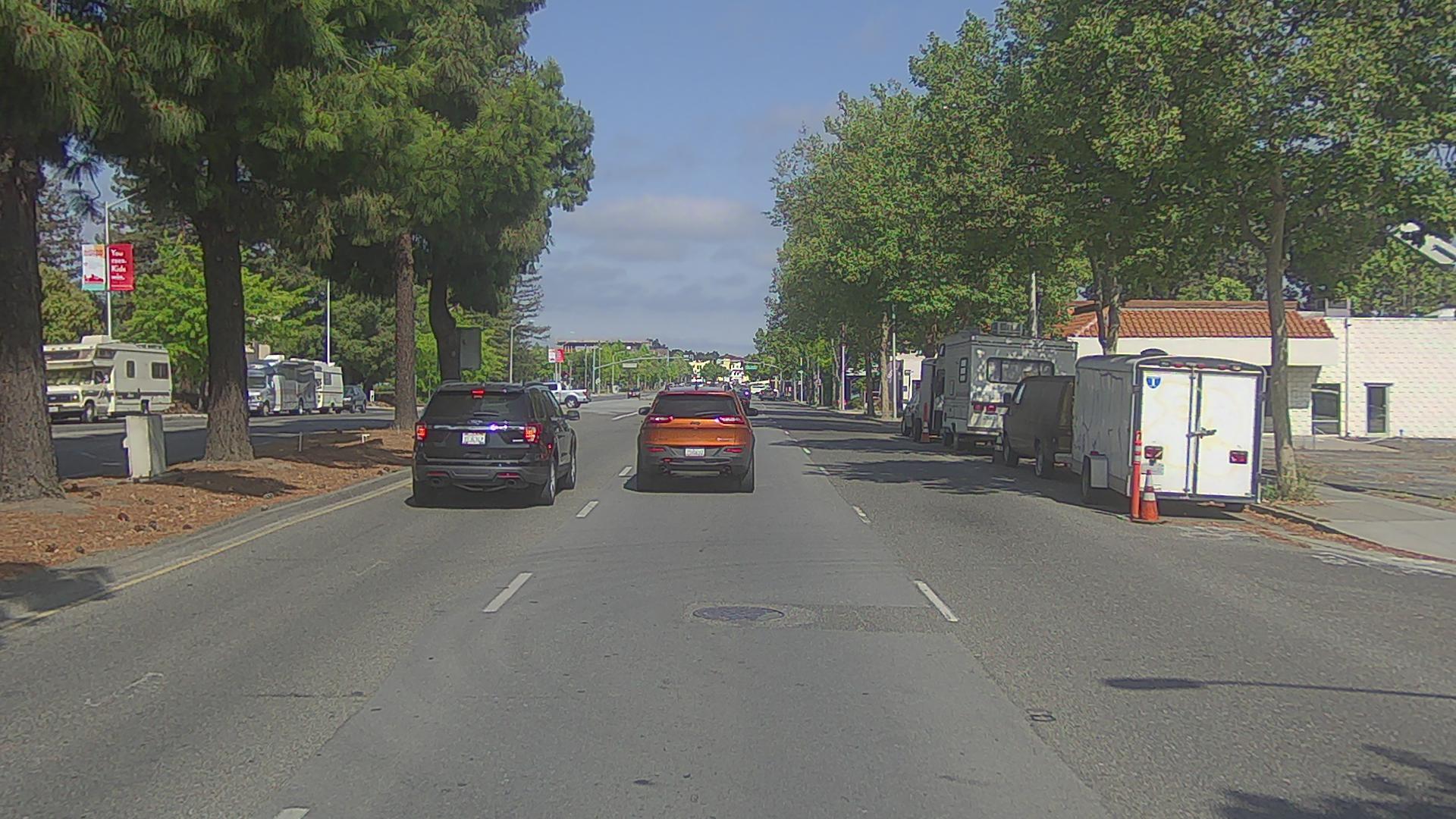}
        \caption{Ground truth}
    \end{subfigure}
\begin{subfigure}{0.8\textwidth}
    \begin{subfigure}{0.49\textwidth}
        \centering
        \includegraphics[width=0.49\linewidth]{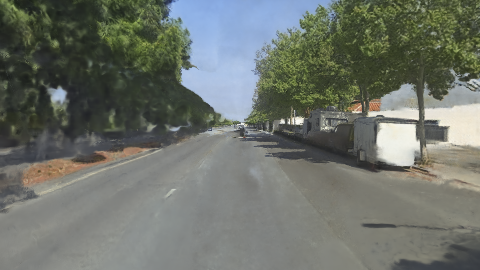}
        \includegraphics[width=0.49\linewidth]{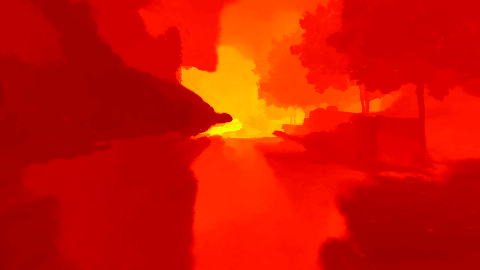}
        \caption{$\mathcal{L}_{robust}$}
    \end{subfigure}
    \hfill
    \begin{subfigure}{0.49\textwidth}
        \centering
        \includegraphics[width=0.49\linewidth]{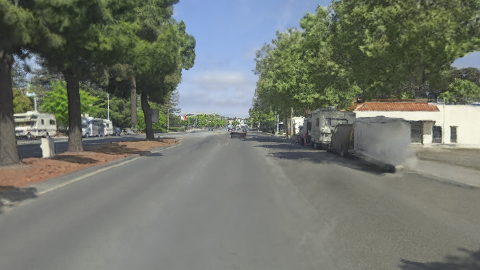}
        \includegraphics[width=0.49\linewidth]{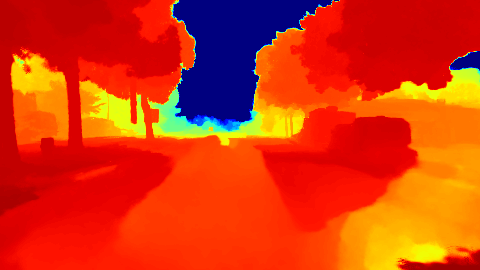}
        \caption{$\mathcal{L}_{robust} + \mathcal{L}_{sky}$}
    \end{subfigure}
    \\
    \begin{subfigure}{0.49\textwidth}
        \centering
        \includegraphics[width=0.49\linewidth]{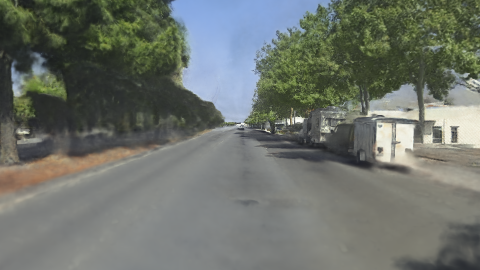}
        \includegraphics[width=0.49\linewidth]{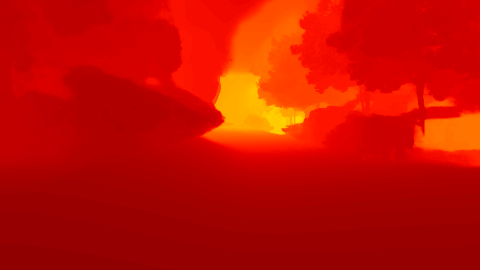}
        \caption{$\mathcal{L}_{robust} + \mathcal{L}_{road}$}
    \end{subfigure}
    \hfill
    \begin{subfigure}{0.49\textwidth}
        \centering
        \includegraphics[width=0.49\linewidth]{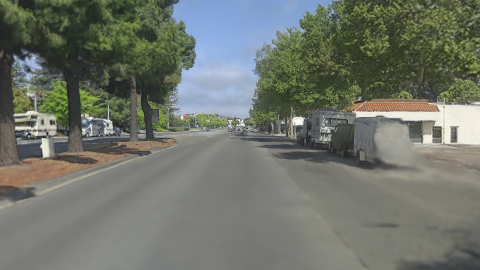}
        \includegraphics[width=0.49\linewidth]{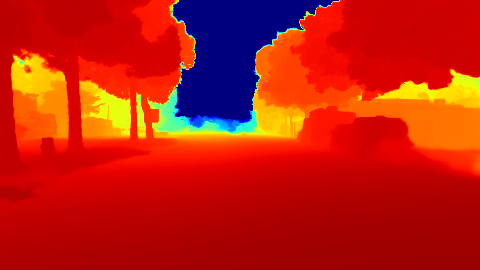}
        \caption{$\mathcal{L}_{robust} + \mathcal{L}_{sky} + \mathcal{L}_{road}$}
    \end{subfigure}
\end{subfigure}
    \caption{(b) Using a trimmed kernel effectively removes all moving cars from the scene. (c, d, e) As the sky and road regions are badly reconstructed, applying class constraints can aid the learning process of the static model.}
    \label{fig:robust_ablation}
\end{figure}

\noindent\textbf{Sky loss.} 
Outdoor scenes contain sky regions where rays never intersect any opaque surfaces. We detect pixels that belong to the sky and force the densities to be zeros along the ray, excluding the last sample:
\begin{equation}
    \mathcal{L}_{sky} = \mathbb{E}\left[\frac{1}{K-1}\sum_{i=1}^{K-1}\sigma_i + \frac{1}{\sigma_K}\right]_{\mathbf{r}\in\mathcal{R}_{sky}}.
    \label{eq:sky_loss}
\end{equation}

\noindent\textbf{Planar regularization loss.}
We observe that the road surface is challenging to model as it has low textures and is often occluded by moving vehicles. These areas mostly suffer from poor geometry prediction (\cref{fig:robust_ablation}). To address this issue, we rely on the work of Wang \etal~\cite{wang2023planerf}, perform Singular Values Decomposition (SVD) on the set of centered points projected from patches that are assigned with the road label, and minimize the smallest singular value $\sigma_3$:
\begin{equation}
    \mathcal{L}_{road} = \sigma_3(\mathcal{R}_{road}^{patch}).
\end{equation}

\noindent\textbf{Dynamic regularization loss.} 
We use the factorization $\mathcal{L}_{\mathcal{H}}$ and sparsity $\mathcal{L}_{\sigma^D}, \mathcal{L}_\rho$ losses that have been proposed in~\cite{yuan2021star, wu2022d, sharma2023neural} to enforce the static branch to explain most of the static structure as possible and only express the moving objects and/or shadows using the dynamic branch.

\noindent Our final loss results on the weighted sum of all our individual losses: 
\begin{equation}
\begin{split}
    \mathcal{L}_{total} &= \underbrace{\mathcal{L}_{rgb} + \lambda_{d}\mathcal{L}_{d} +\lambda_{sem}\mathcal{L}_{sem} }_{reconstruction}\\
    &+ \underbrace{\mathcal{L}_{robust} + \lambda_{sky}\mathcal{L}_{sky} + \lambda_{road}\mathcal{L}_{road}}_{static\ regularization} 
    + \underbrace{\lambda_{\sigma^D}\mathcal{L}_{\sigma^D} + \lambda_{\rho}\mathcal{L}_{\rho} + \lambda_{\mathcal{H}}\mathcal{L}_{\mathcal{H}}}_{dynamic\ regularization}
\end{split}
\end{equation}

\begin{figure}[tp]
\begin{minipage}[b]{0.03\textwidth}
     \rotatebox{90}{GT}
\end{minipage}
\hfill
\raisebox{-1\baselineskip}{
\begin{subfigure}[b]{0.32\linewidth}
    \includegraphics[width=\linewidth]{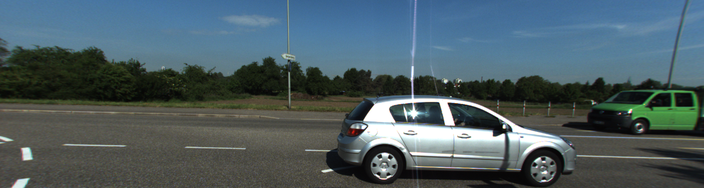}
\end{subfigure}
\begin{subfigure}[b]{0.32\linewidth}
    \includegraphics[width=\linewidth]{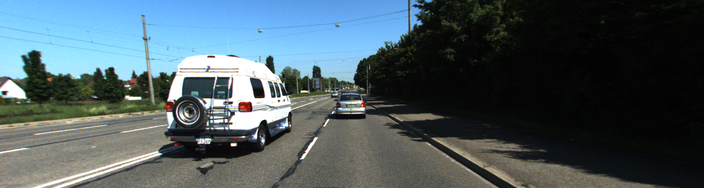}
\end{subfigure}
\begin{subfigure}[b]{0.32\linewidth}
    \includegraphics[width=\linewidth]{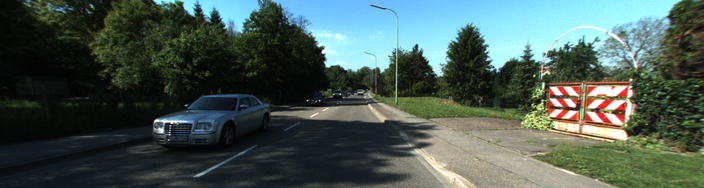}
\end{subfigure}}

\begin{minipage}[b]{0.03\textwidth}
     \rotatebox{90}{$\hat{I}$}
\end{minipage}
\hfill
\raisebox{-1\baselineskip}{
\begin{subfigure}[b]{0.32\linewidth}
    \includegraphics[width=\linewidth]{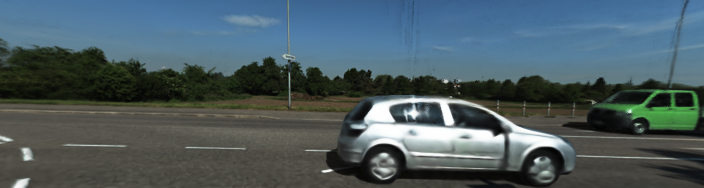}
\end{subfigure}
\begin{subfigure}[b]{0.32\linewidth}
    \includegraphics[width=\linewidth]{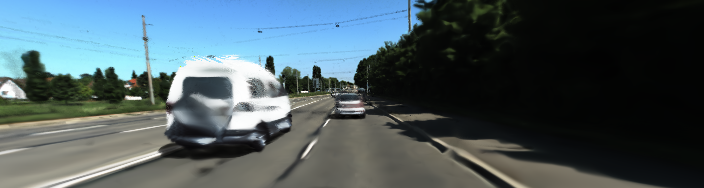}
\end{subfigure}
\begin{subfigure}[b]{0.32\linewidth}
    \includegraphics[width=\linewidth]{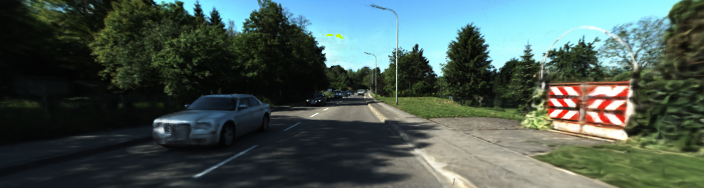}
\end{subfigure}}

\begin{minipage}[b]{0.03\textwidth}
     \rotatebox{90}{$\hat{d}$}
\end{minipage}
\hfill
\raisebox{-1\baselineskip}{
\begin{subfigure}[b]{0.32\linewidth}
    \includegraphics[width=\linewidth]{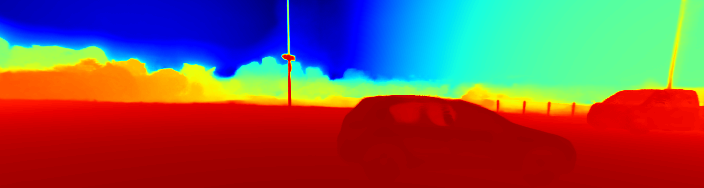}
\end{subfigure}
\begin{subfigure}[b]{0.32\linewidth}
    \includegraphics[width=\linewidth]{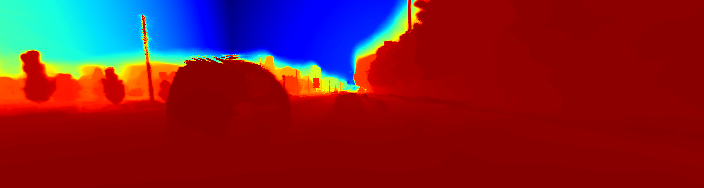}
\end{subfigure}
\begin{subfigure}[b]{0.32\linewidth}
    \includegraphics[width=\linewidth]{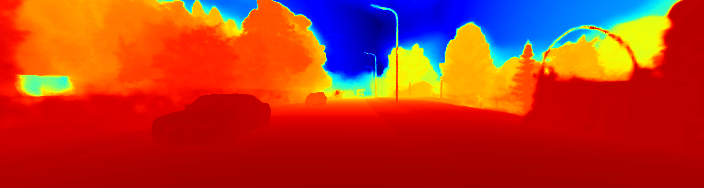}
\end{subfigure}}

\begin{minipage}[b]{0.03\textwidth}
     \rotatebox{90}{$\hat{I}^S$}
\end{minipage}
\hfill
\raisebox{-1\baselineskip}{
\begin{subfigure}[b]{0.32\linewidth}
    \includegraphics[width=\linewidth]{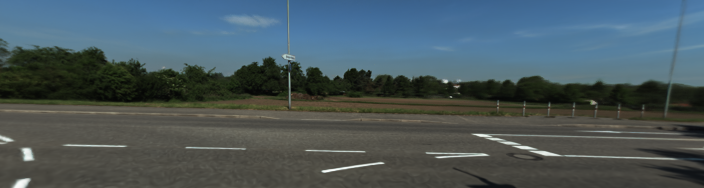}
\end{subfigure}
\begin{subfigure}[b]{0.32\linewidth}
    \includegraphics[width=\linewidth]{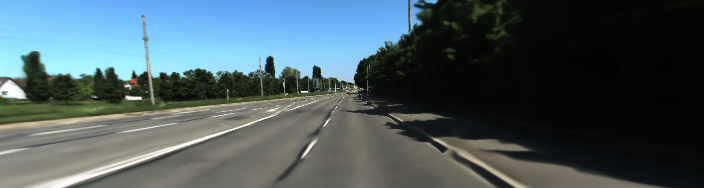}
\end{subfigure}
\begin{subfigure}[b]{0.32\linewidth}
    \includegraphics[width=\linewidth]{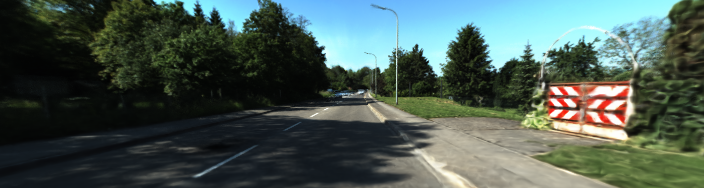}
\end{subfigure}}

\begin{minipage}[b]{0.03\textwidth}
     \rotatebox{90}{$\hat{d}^S$}
\end{minipage}
\hfill
\raisebox{-1\baselineskip}{
\begin{subfigure}[b]{0.32\linewidth}
    \includegraphics[width=\linewidth]{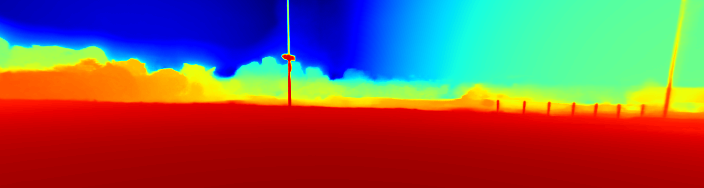}
\end{subfigure}
\begin{subfigure}[b]{0.32\linewidth}
    \includegraphics[width=\linewidth]{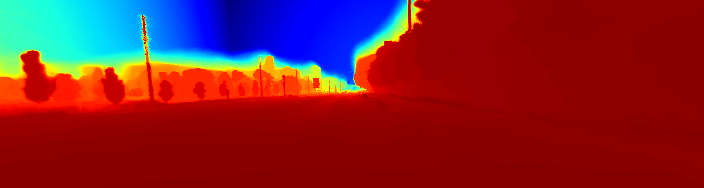}
\end{subfigure}
\begin{subfigure}[b]{0.32\linewidth}
    \includegraphics[width=\linewidth]{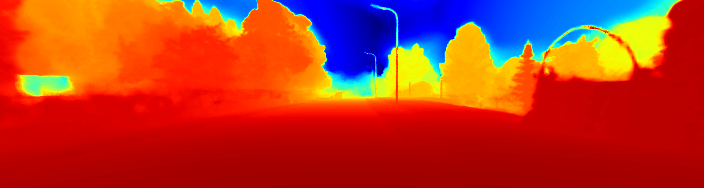}
\end{subfigure}}

\begin{minipage}[b]{0.03\textwidth}
     \rotatebox{90}{$\hat{I}^D$}
\end{minipage}
\hfill
\raisebox{-1\baselineskip}{
\begin{subfigure}[b]{0.32\linewidth}
    \includegraphics[width=\linewidth]{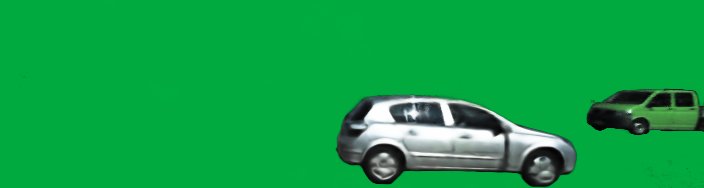}
\end{subfigure}
\begin{subfigure}[b]{0.32\linewidth}
    \includegraphics[width=\linewidth]{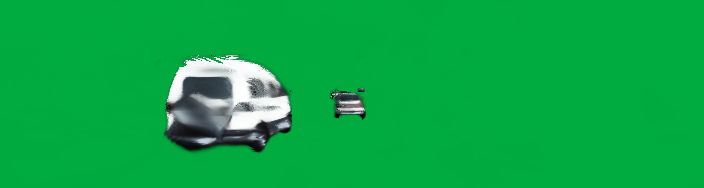}
\end{subfigure}
\begin{subfigure}[b]{0.32\linewidth}
    \includegraphics[width=\linewidth]{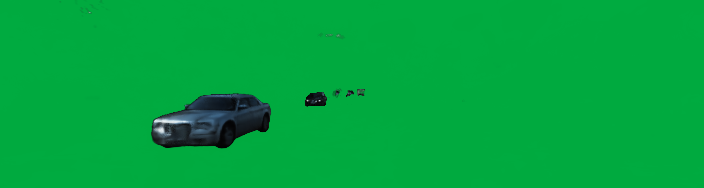}
\end{subfigure}}

\begin{minipage}[b]{0.03\textwidth}
     \rotatebox{90}{$\hat{S}$}
\end{minipage}
\hfill
\raisebox{-1\baselineskip}{
\begin{subfigure}[b]{0.32\linewidth}
    \includegraphics[width=\linewidth]{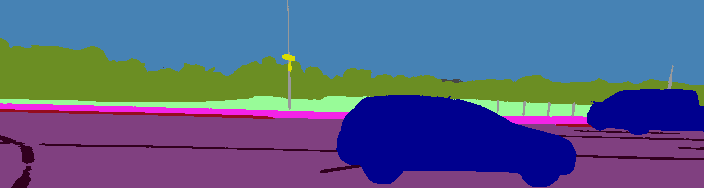}
\end{subfigure}
\begin{subfigure}[b]{0.32\linewidth}
    \includegraphics[width=\linewidth]{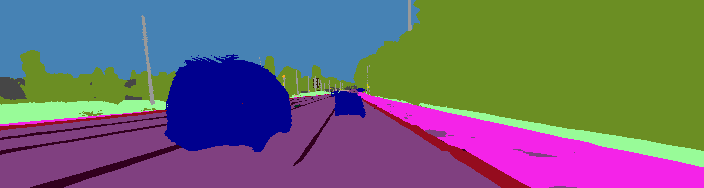}
\end{subfigure}
\begin{subfigure}[b]{0.32\linewidth}
    \includegraphics[width=\linewidth]{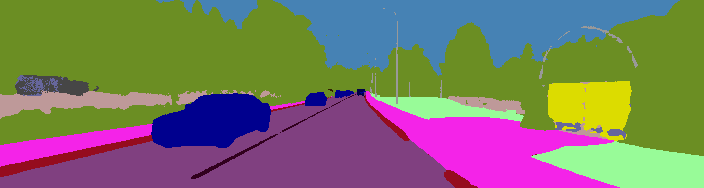}
\end{subfigure}}

\begin{minipage}[b]{0.03\textwidth}
     \rotatebox{90}{$\hat{S}^S$}
\end{minipage}
\hfill
\raisebox{-1\baselineskip}{
\begin{subfigure}[b]{0.32\linewidth}
    \includegraphics[width=\linewidth]{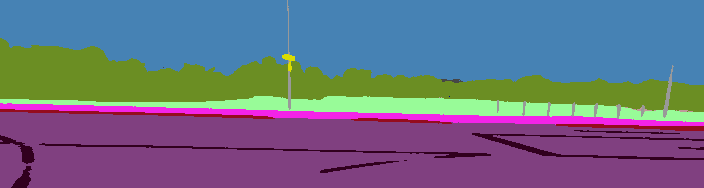}
\end{subfigure}
\begin{subfigure}[b]{0.32\linewidth}
    \includegraphics[width=\linewidth]{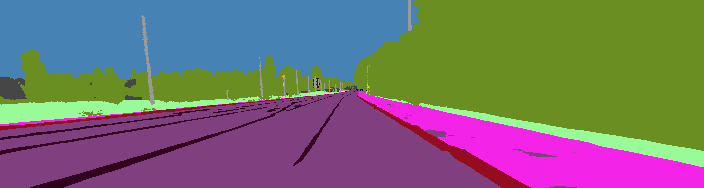}
\end{subfigure}
\begin{subfigure}[b]{0.32\linewidth}
    \includegraphics[width=\linewidth]{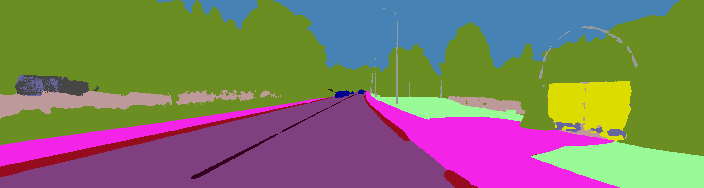}
\end{subfigure}}

\begin{minipage}[b]{0.03\textwidth}
     \rotatebox{90}{$M^D$}
\end{minipage}
\hfill
\raisebox{-1\baselineskip}{
\begin{subfigure}[b]{0.32\linewidth}
    \includegraphics[width=\linewidth]{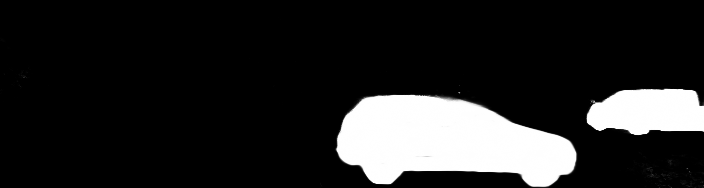}
\end{subfigure}
\begin{subfigure}[b]{0.32\linewidth}
    \includegraphics[width=\linewidth]{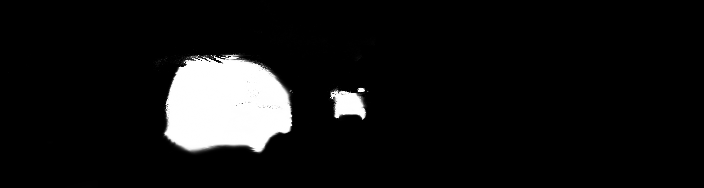}
\end{subfigure}
\begin{subfigure}[b]{0.32\linewidth}
    \includegraphics[width=\linewidth]{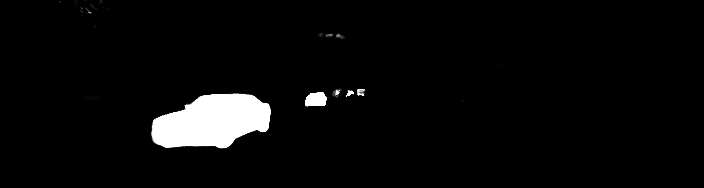}
\end{subfigure}}
    \caption{We demonstrate RoDUS's capabilities on multiple tasks. \textit{Legend:} GT: Ground truth, $\hat{I}$: Composed RGB, $\hat{d}$: Composed depth, $I^S$: Static RGB, $\hat{d}^S$: Static depth, $\hat{I}^D$: Dynamic RGB (blended with a green background), $\hat{S}$: Composed segmentation map, $\hat{S}^S$: Static segmentation map, $M^D$: Motion segmentation.}
    \label{fig:qualitative}
\end{figure}

\subsection{Training Strategy}
\label{sec:strategy}
The use of robust estimation plays an important role in rejecting inconsistency (dynamic objects in our case) and guiding the static branch without any prior knowledge. However, we observe that when using a binary kernel alone with a fixed threshold, after reaching a certain point throughout the training, the robust loss does not contribute to the overall loss anymore. 
We refer to this phenomenon as ``kernel saturation'', this causes several parts of the scene not to be learned completely. Adjusting the threshold $\mathcal{T}_\epsilon$ seems to be a trivial solution, but given the fact that outdoor scenes are more complicated to handle, there is always a trade-off between the number of distractors removed and the static scene quality. 
Particularly, high-frequency but static details, equivalently appear as large residuals are considered as outliers, while low-frequency but moving regions such as shadows (\textbf{dark}) which mostly share a similar color with the road \textcolor{gray}{(\textbf{gray})} are uncontrollably included as inliers. 
\cref{fig:qualitative_panda} portrays this, as there are still parts of the scene (\eg road marking) that are not fully learned.

In contrast, relying solely on reconstruction losses leads to several local optima during optimization due to the entanglement of geometry and appearance. 
We propose using the robust loss as an initialization step for the static model to learn the basic overall structures of the scene before delving into complete scene understanding. 
The initialization step guides the network to prioritize consistent (or confident) regions that further help stabilize the training process. Therefore, we start the model training with the robust loss (\cref{eq:robust_loss}) and sky loss (\cref{eq:sky_loss}) enabled for the first epoch, and subsequently enable the remaining losses.
By doing so, we can confidently select a low threshold without compromising the quality. This serves as a foundation for disentangling the scene in later training stages.

\section{Evaluation}
\label{sec:evaluation}

\subsection{Experimental Setup}
\textbf{Implementation details.} We build RoDUS using PyTorch~\cite{paszke2019pytorch}. Our implementation is based on \texttt{tiny-cuda-nn}~\cite{muller2021tinycudann} from iNGP~\cite{muller2022instant}, and proposal sampling strategy inspired from MipNeRF360~\cite{barron2022mip}. 
To generate pseudo-ground truth for semantic segmentation, we use the Mask2Former~\cite{cheng2022masked} model pre-trained on the Mapillary Vistas~\cite{neuhold2017mapillary} dataset.

\noindent\textbf{Datasets.} We evaluate our system on two challenging urban driving datasets, KITTI-360~\cite{liao2022kitti} and Pandaset~\cite{xiao2021pandaset}, both featuring dynamic objects in driving scenarios. In Pandaset, we utilize three front cameras and one back camera, while in KITTI-360, we use all two perspective cameras along with side fish-eye cameras.
For evaluation, we rely on provided ground-truth instance masks (KITTI-360)/projected 3D bounding boxes (Pandaset) with motion labels to identify true dynamic objects. We refer to the supplementary for additional details on the selection of the training sequences and data pre-processing.

\noindent\textbf{Baseline.}
We benchmark RoDUS against 4 different state-of-the-art methods for static scene extraction (RobustNeRF~\cite{sabour2023robustnerf}) and dynamic scene rendering (D$^2$NeRF~\cite{wu2022d}, SUDS~\cite{turki2023suds}, and EmerNeRF~\cite{yang2023emernerf}). Evaluations focus on two key factors: the quality of novel view synthesis of the static image and the segmentation masks of dynamic objects. While our primary focus is on static-dynamic disentanglement, RoDUS's design allows it to represent dynamic scenes holistically so the quality of the overall image is also considered. We showcase the performance of our method on three representative scenarios in \cref{fig:qualitative}

\begin{table}[tb]
    
    \caption{Quantitative comparisons with SoTA methods on KITTI-360 and Pandaset.
    }
    \label{tab:headings}
    \centering
    \resizebox{\columnwidth}{!}{%
  \begin{tabular}{lclccccccccc}
    \cmidrule[\heavyrulewidth]{4-12}
     & & & \multicolumn{3}{c}{Image Reconstruction} & \multicolumn{3}{c}{Static NVS} & \multicolumn{3}{c}{Motion segmentation}
    \\
    \cmidrule(rl){4-6} 
    \cmidrule(rl){7-9}
    \cmidrule(rl){10-12}

    & & & PSNR$\uparrow$ & SSIM$\uparrow$ & LPIPS$\downarrow$ & PSNR$\uparrow$ & SSIM$\uparrow$ & LPIPS$\downarrow$ & Recall$\uparrow$ & IoU$\uparrow$ & F1$\uparrow$\\
    \midrule
    
    \multirow{6}{*}{\rotatebox[origin=c]{90}{~~~~KITTI-360}} & & RobustNeRF~\cite{sabour2023robustnerf}$^{\dag}$ & - & - & - & 19.50 & 0.722 & 0.282 & - & - & -\\
    & & D$^2$NeRF~\cite{wu2022d} & 23.86 & 0.688 & 0.232 & 18.70 & 0.588 & 0.309& 12.04 & 4.67 & 7.41\\
    &  & SUDS~\cite{turki2023suds} & \textbf{28.16} & \underline{0.836} & \underline{0.175} & 18.81 & 0.687 & 0.298 & \textbf{93.89} & 5.35 & 9.34\\
    &  & EmerNeRF~\cite{yang2023emernerf} & 25.49 & 0.798 & 0.181 & \underline{20.59} & \underline{0.733} & \underline{0.249} & 57.74 & \underline{12.56} & \underline{19.92}\\
    &  & RoDUS (ours) & \underline{26.47} & \textbf{0.857} & \textbf{0.148} & \textbf{23.43} & \textbf{0.802} & \textbf{0.181} & \underline{79.45} & \textbf{64.84} & \textbf{76.52}\\
    \midrule
    
    \multirow{6}{*}{\rotatebox[origin=c]{90}{~~~~Pandaset}} & ~~ &  RobustNeRF~\cite{sabour2023robustnerf}$^{\dag}$ &    - & - & - & 23.41 & 0.764 & 0.197 & - & - & -\\
    &  & D$^2$NeRF~\cite{wu2022d}  & 25.25 & 0.692 & 0.291 & 23.38 & 0.682 & 0.292 & - & - & -\\
    &  & SUDS~\cite{turki2023suds}  & \textbf{30.28} & \underline{0.851} & \underline{0.138} & 25.34 & 0.788 & 0.182 & - & - & -\\
    &  &EmerNeRF~\cite{yang2023emernerf}  & \underline{29.51} & 0.841 & 0.155 & \textbf{27.47} & \underline{0.811} & \underline{0.166} & - & - & -\\
    &  &RoDUS (ours)  &  28.39 & \textbf{0.877} & \textbf{0.135} & \underline{25.83} & \textbf{0.825} & \textbf{0.156} & - & - & -\\
    \bottomrule
    \\
  \end{tabular}
    }
    
    \vspace{-0.2cm}
    {\tiny $^{\dag}$ As RobustNeRF rejects distractors during training, full image reconstruction metrics are not included.}
    
    \label{tab:reconstruction}  
\end{table}

\begin{figure*}[tp]
\centering
\scriptsize
\setlength{\tabcolsep}{0.002\linewidth}
\renewcommand{\arraystretch}{0.8}
\begin{tabular}{ccccc} 
    \multirow{1}{*}[10mm]{\rotatebox[origin=c]{90}{\scriptsize GT}} & &
    \includegraphics[width=0.30\textwidth]{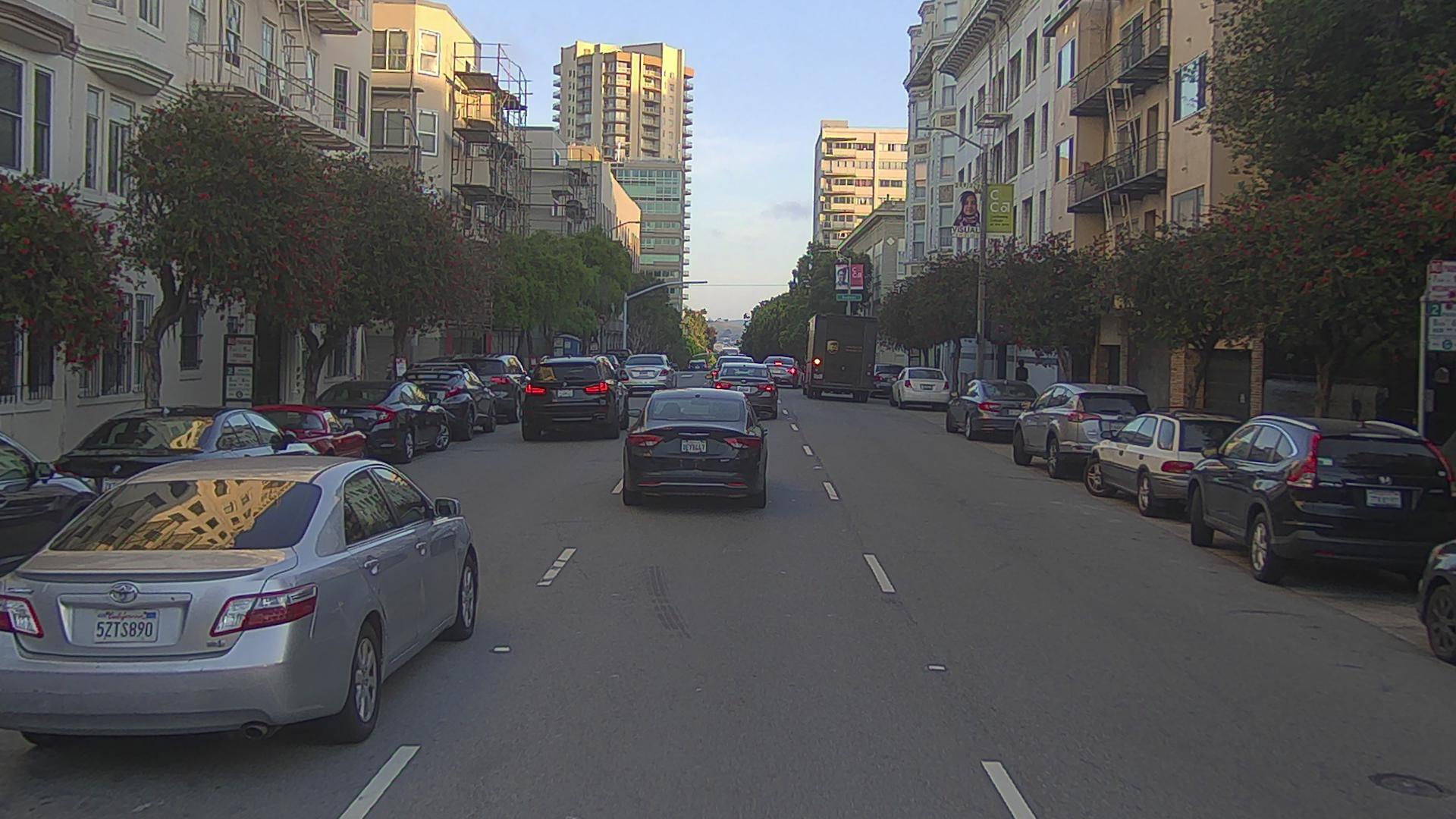} &  & \\

    \multirow{1}{*}[16mm]{\rotatebox[origin=c]{90}{\scriptsize RobustNeRF}} &
    \multirow{1}{*}[10mm]{\rotatebox[origin=c]{90}{\scriptsize \cite{sabour2023robustnerf}}} &
    \framebox(0.29\linewidth,0.165\linewidth){\small N/A} & 
    \includegraphics[width=0.30\textwidth]{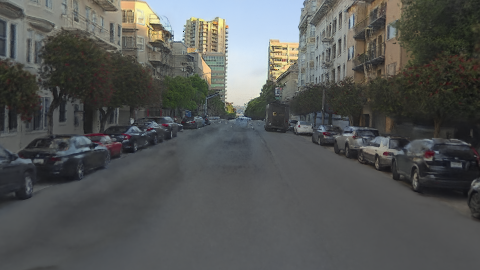} & 
    \framebox(0.29\linewidth,0.165\linewidth){\small N/A} \\

    \multirow{1}{*}[12mm]{\rotatebox[origin=c]{90}{\scriptsize D$^2$NeRF}} &
    \multirow{1}{*}[10mm]{\rotatebox[origin=c]{90}{\scriptsize \cite{wu2022d}}} &
    \includegraphics[width=0.30\textwidth]{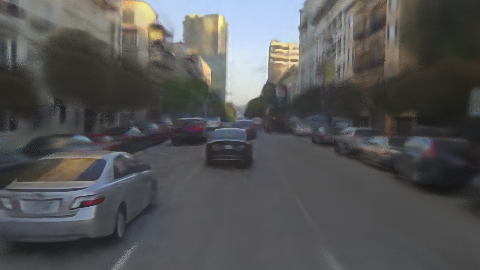} & 
    \includegraphics[width=0.30\textwidth]{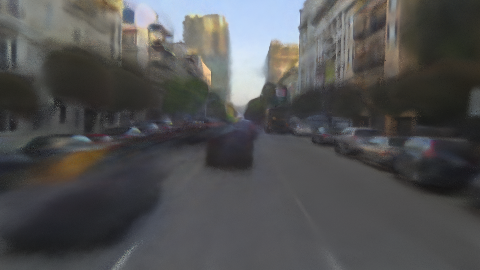} & 
    \includegraphics[width=0.30\textwidth]{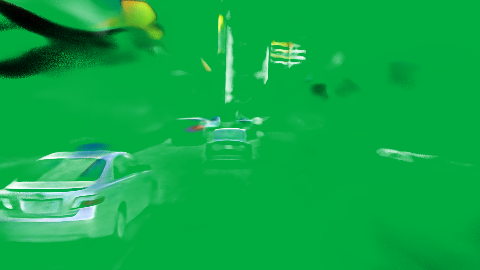} \\

    \multirow{1}{*}[12mm]{\rotatebox[origin=c]{90}{\scriptsize SUDS}} &
    \multirow{1}{*}[10mm]{\rotatebox[origin=c]{90}{\scriptsize \cite{turki2023suds}}} &
    \includegraphics[width=0.30\textwidth]{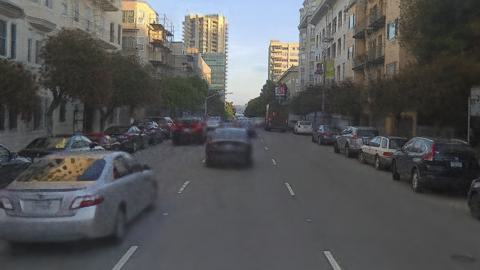} & 
    \includegraphics[width=0.30\textwidth]{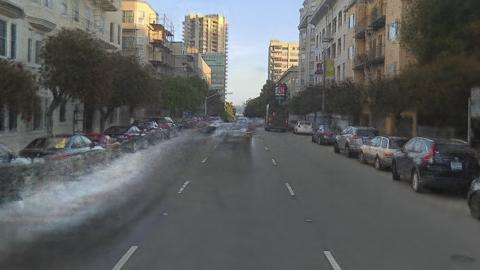} & 
    \includegraphics[width=0.30\textwidth]{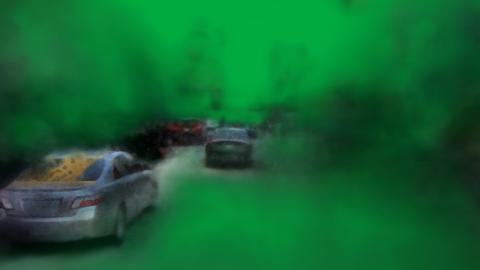} \\

    \multirow{1}{*}[14mm]{\rotatebox[origin=c]{90}{\scriptsize EmerNeRF}} &
    \multirow{1}{*}[10mm]{\rotatebox[origin=c]{90}{\scriptsize \cite{yang2023emernerf}}} &
    \includegraphics[width=0.30\textwidth]{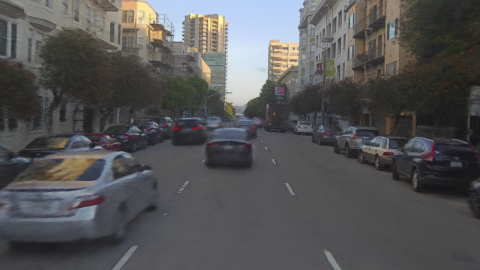} & 
    \includegraphics[width=0.30\textwidth]{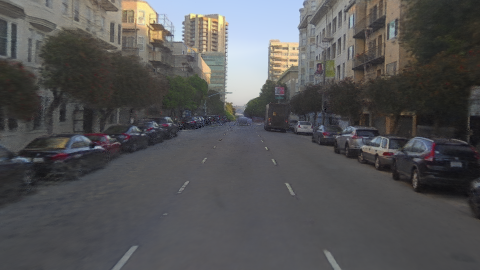} & 
    \includegraphics[width=0.30\textwidth]{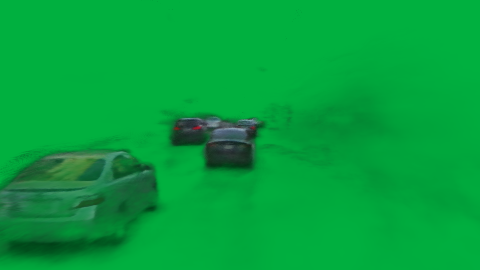} \\

    \multirow{1}{*}[12mm]{\rotatebox[origin=c]{90}{\scriptsize RoDUS}} &
    \multirow{1}{*}[12mm]{\rotatebox[origin=c]{90}{\scriptsize (ours)}} &
    \includegraphics[width=0.30\textwidth]{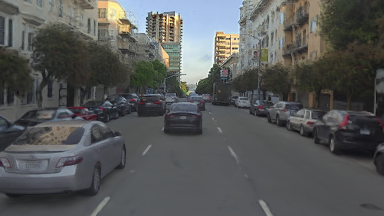} & 
    \includegraphics[width=0.30\textwidth]{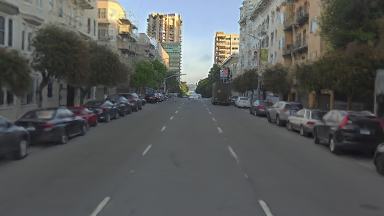} & 
    \includegraphics[width=0.30\textwidth]{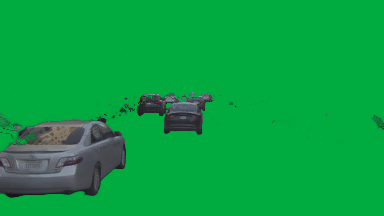} \\           
    & ~ & \small Composed RGB & \small Static RGB & \small Dynamic RGB \\
            
\end{tabular}
\caption{Reconstruction and decomposition qualitative results comparisons on Pandaset.
D$^2$NeRF struggles in intricate driving contexts and produces poor results. SUDS faces difficulties in distinguishing between ego-motion and object-motion due to its reliance on optical flow supervision, leading to a large number of cloudy artifacts in the dynamic image to match the ground truth optical flows. RobustNeRF successfully reconstructs the background with no dynamic objects remaining, but it suffers from saturation, leading to degraded results (as explained in \cref{sec:loss}). We can see that the separation quality is still suboptimal. In contrast, RoDUS successfully decouples dynamic objects and achieves a clean background with no artifacts remaining.
}
\label{fig:qualitative_panda}
\end{figure*}

\subsection{Static-Dynamic Decomposition}
We report PSNR, SSIM~\cite{wang2004image}, and LPIPS~\cite{zhang2018unreasonable} for both image reconstruction and static NVS tasks.
We create a validation set by splitting every 8$^{\text{th}}$ image (sorted by timestamps) per camera. In the image reconstruction task, we utilize all available samples of the training set for evaluation. In the static NVS task, we compare the image rendered from the static branch with the ground truth with all true dynamic objects masked out. 
As illustrated in \cref{fig:qualitative_panda} and \cref{tab:reconstruction}, other methods achieve good rendering quality in the composed image. However, only EmerNeRF and RoDUS achieve reasonable results when rendering each component separately. We also lead in SSIM and LPIPS.

\subsection{Dynamic Objects Segmentation}
\begin{figure*}[tp]
	\centering
	\scriptsize
	\setlength{\tabcolsep}{0.002\linewidth}
	\renewcommand{\arraystretch}{0.8}
	\begin{tabular}{ccccc}
        
        \multirow{1}{*}[5.7mm]{\rotatebox[origin=c]{90}{\scriptsize GT}} & &
        \includegraphics[clip=true, trim={0 0 100 0},width=0.32\textwidth]{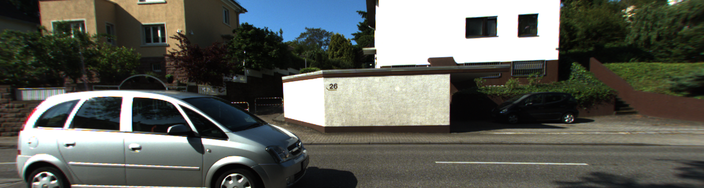} & 
        \includegraphics[clip=true, trim={50 0 50 0},width=0.32\textwidth]{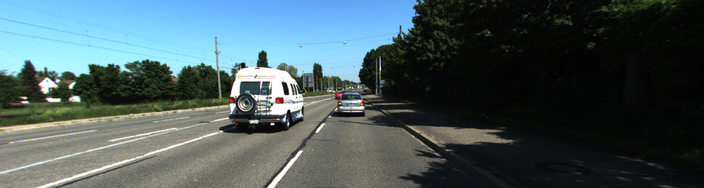} & 
        \includegraphics[clip=true, trim={50 0 50 0},width=0.32\textwidth]{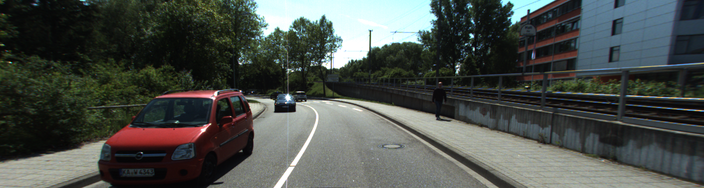}
        \\

        \multirow{1}{*}[5.7mm]{\rotatebox[origin=c]{90}{\scriptsize GT}} &
        \multirow{1}{*}[6.5mm]{\rotatebox[origin=c]{90}{\scriptsize mask}} &
        \includegraphics[clip=true, trim={0 0 100 0},width=0.32\textwidth]{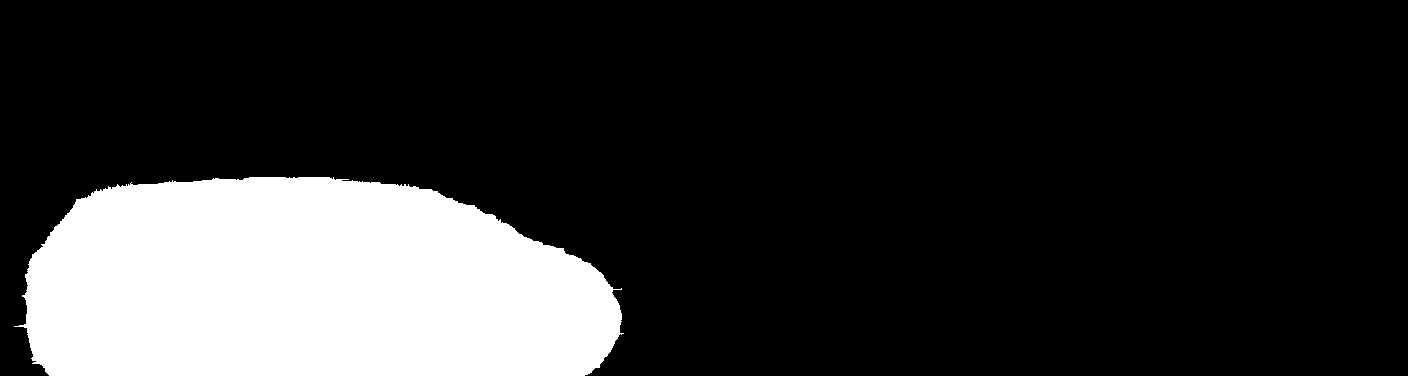} & 
        \includegraphics[clip=true, trim={50 0 50 0},width=0.32\textwidth]{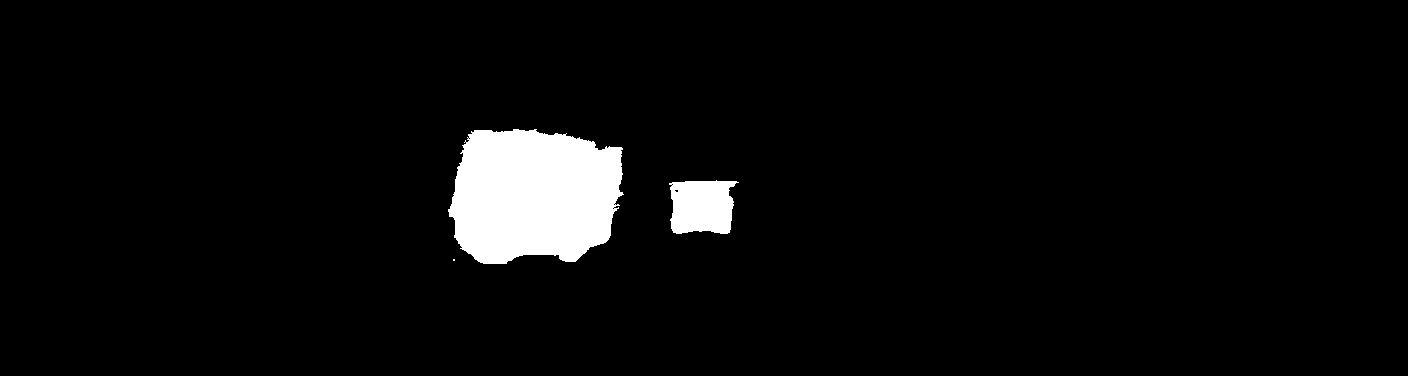} & 
        \includegraphics[clip=true, trim={50 0 50 0},width=0.32\textwidth]{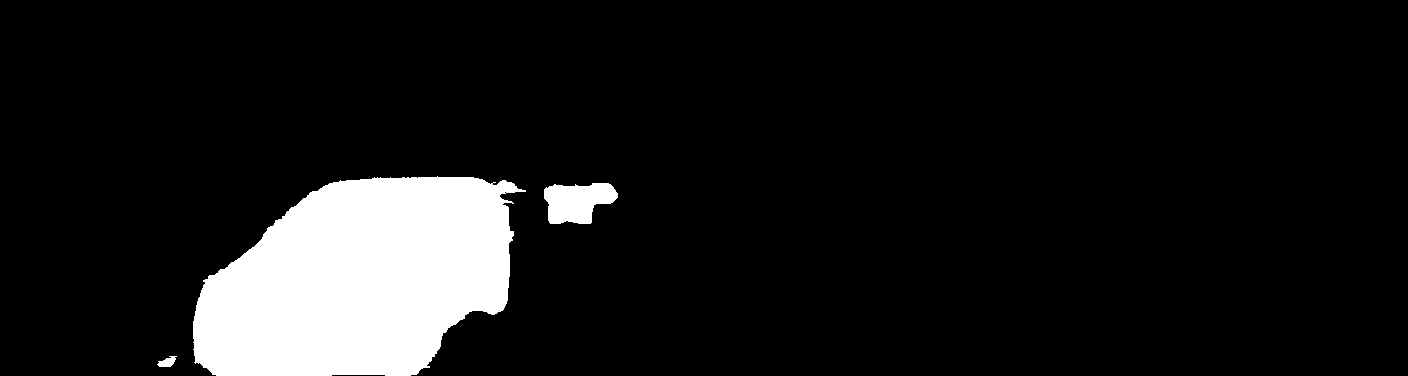}
        \\

        \multirow{1}{*}[9mm]{\rotatebox[origin=c]{90}{\scriptsize D$^2$NeRF}} &
        \multirow{1}{*}[6mm]{\rotatebox[origin=c]{90}{\scriptsize \cite{wu2022d}}} &
        \includegraphics[clip=true, trim={0 0 100 0},width=0.32\textwidth]{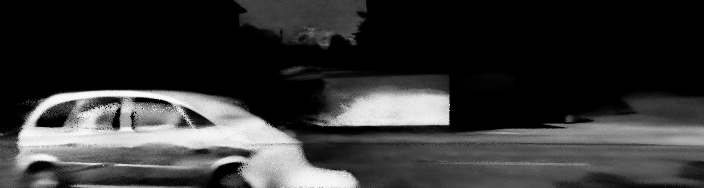} & 
        \includegraphics[clip=true, trim={50 0 50 0},width=0.32\textwidth]{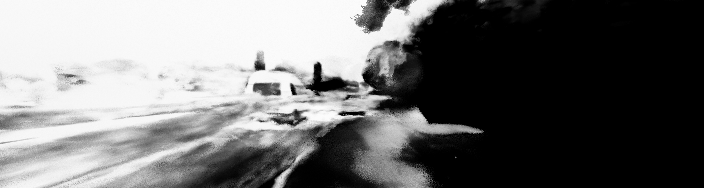} & 
        \includegraphics[clip=true, trim={50 0 50 0},width=0.32\textwidth]{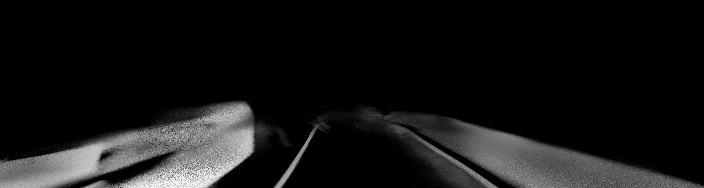}
        \\

        \multirow{1}{*}[7.5mm]{\rotatebox[origin=c]{90}{\scriptsize SUDS}} &
        \multirow{1}{*}[6mm]{\rotatebox[origin=c]{90}{\scriptsize \cite{turki2023suds}}} &
        \includegraphics[clip=true, trim={0 0 100 0},width=0.32\textwidth]{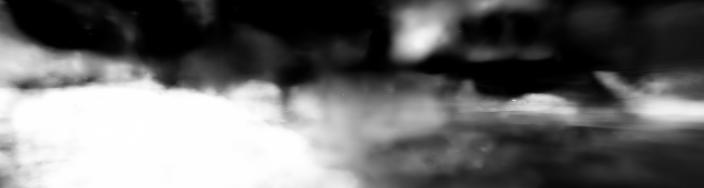} & 
        \includegraphics[clip=true, trim={50 0 50 0},width=0.32\textwidth]{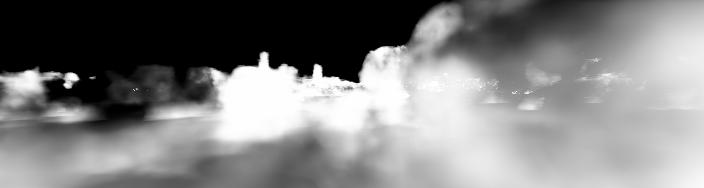} & 
        \includegraphics[clip=true, trim={50 0 50 0},width=0.32\textwidth]{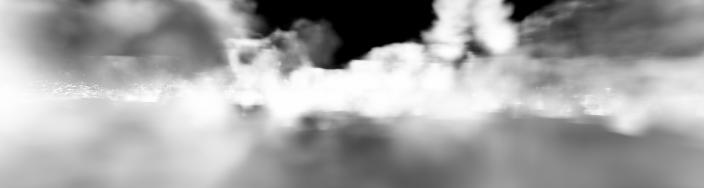}
        \\

        \multirow{1}{*}[10mm]{\rotatebox[origin=c]{90}{\scriptsize EmerNeRF}} &
        \multirow{1}{*}[6mm]{\rotatebox[origin=c]{90}{\scriptsize \cite{yang2023emernerf}}} &
        \includegraphics[clip=true, trim={0 0 100 0},width=0.32\textwidth]{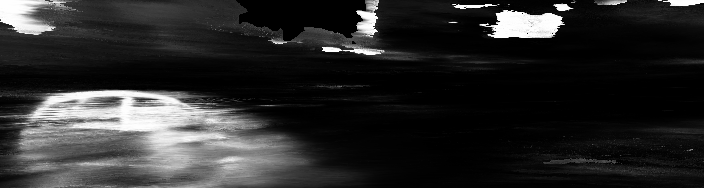} & 
        \includegraphics[clip=true, trim={50 0 50 0},width=0.32\textwidth]{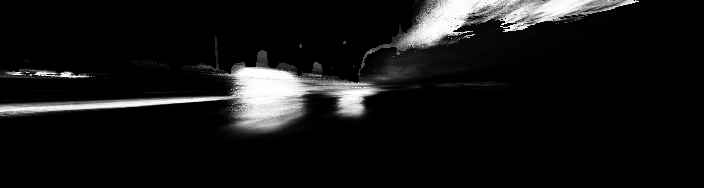} & 
        \includegraphics[clip=true, trim={50 0 50 0},width=0.32\textwidth]{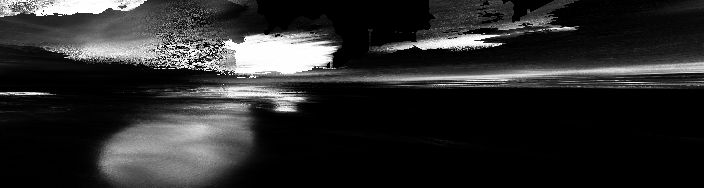}
        \\

        \multirow{1}{*}[8mm]{\rotatebox[origin=c]{90}{\scriptsize RoDUS}} &
        \multirow{1}{*}[7mm]{\rotatebox[origin=c]{90}{\scriptsize (ours)}} &
        \includegraphics[clip=true, trim={0 0 100 0},width=0.32\textwidth]{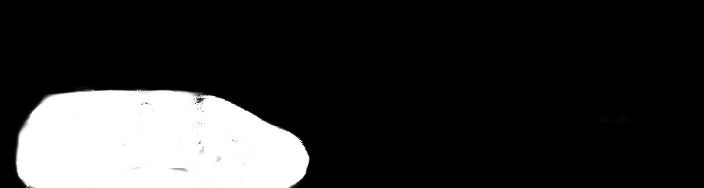} & 
        \includegraphics[clip=true, trim={50 0 50 0},width=0.32\textwidth]{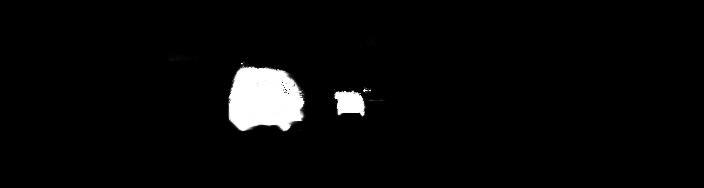} & 
        \includegraphics[clip=true, trim={50 0 50 0},width=0.32\textwidth]{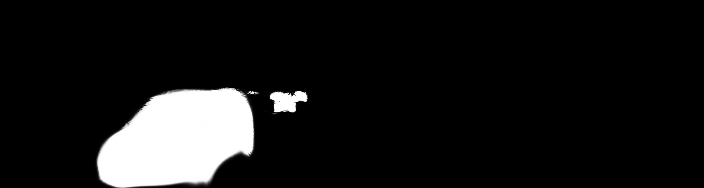}
        \\

        & ~ & (a) & (b) & (c) \\

	\end{tabular}
	\caption{Motion segmentation qualitative results comparison on KITTI-360 before thresholding. We present results in three scenarios: (a) The ego-car stops in front of an intersection; (b) Vehicles move in the same direction as the ego-car; and (c) Vehicles move in the opposite direction toward the ego-car.
    }
	\label{fig:seg}
\end{figure*}

For this task, we obtain the motion mask $M^D$ by rendering dynamic opacities and applying a threshold $\mathcal{T}_{O^D}=0.5$. We report mask Recall, IoU, and F1-score on training views. We only evaluate on KITTI-360 due to the absence of ground-truth instance masks in Pandaset. In \cref{fig:seg}, we explore various scenarios to understand how the ego-actor moves in relation to other moving objects and how it observes the scenes that could impact the decomposition. Particularly, case (a) is likely the easiest as the background is broadly observed for a long period, we can still notice a good separation in D$^2$NeRF, SUDS's results. In cases (b) and (c), when the ego actor is in motion, most methods suffer from incorrect segmentation, while our semantic guidance forces the model to learn the correct dynamic objects. SUDS tends to produce large dynamic areas, resulting in very high Recall but low IoU and F1, yet EmerNeRF exhibits sensitivity to lighting variations. Correspondingly, \cref{tab:reconstruction} indicates that RoDUS's metrics outperform the others in both tasks.

\subsection{Ablation Study}
\begin{figure}
\begin{minipage}[b]{.49\textwidth}
    \captionof{table}{\textbf{Ablation study.} Impact of our design choice on static NVS task.}
    \centering
    \setlength{\tabcolsep}{3pt}
    \begin{tabular}{lccc}
        \toprule
        Ablation & PSNR$\uparrow$ & SSIM$\uparrow$ & LPIPS$\downarrow$ \\
        \midrule
        Full model & 25.94 & 0.814 & 0.170\\
        \midrule
        w/o $\mathcal{L}_d$ & -0.80 & -0.032 & +0.023\\
        w/o $\mathcal{L}_{sem}$ & -1.67	&-0.080&	+0.078\\
        w/o $\mathcal{L}_{robust}$ & -0.91	&-0.047&+0.067\\
        w/o $\mathcal{L}_{road}$ & -0.84& -0.035	&+0.026\\
        w/o $\mathcal{L}_{\sigma^D}$ & -0.74&	-0.030&+0.024\\
        w/o FG mask & -0.67&-0.035&+0.023\\
         \bottomrule
    \end{tabular}
    \label{tab:ablation}
\end{minipage}
\hfill
\begin{minipage}[b]{0.49\textwidth}
    \centering
    \begin{subfigure}{0.49\textwidth}
        \includegraphics[width = \textwidth]{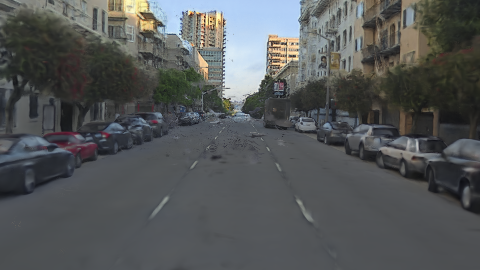}
        \caption*{w/o FG mask}
    \end{subfigure}
    \begin{subfigure}{0.49\textwidth}
        \includegraphics[width = \textwidth]{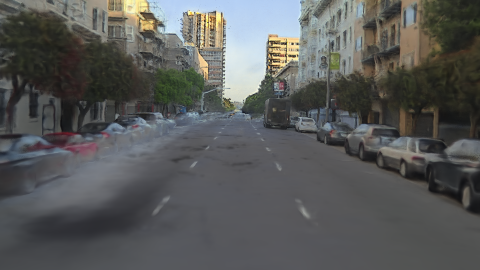}
        \caption*{w/o $\mathcal{L}_{sem}$}
    \end{subfigure}
    \begin{subfigure}{0.49\textwidth}
        \includegraphics[width = \textwidth]{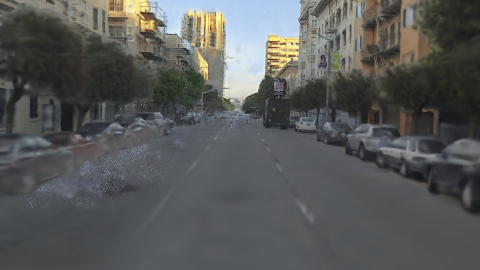}
        \caption*{w/o $\mathcal{L}_{robust}$}
    \end{subfigure}
    \begin{subfigure}{0.49\textwidth}
        \includegraphics[width = \textwidth]{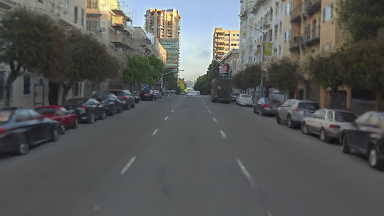}
        \caption*{Full model}
    \end{subfigure}
    \vspace{-0.2cm}
    \captionof{figure}{Components ablation on $\hat{I}^S$.}
    \label{fig:depth_ablation}
\end{minipage}
\end{figure}

We analyze our design choices in \cref{tab:ablation}. The use of road, sky regularization, and robust initialization terms support the geometry, especially in regions occluded by dynamic objects, thus improving the overall quality. Please refer to \cref{fig:robust_ablation} and \cref{fig:depth_ablation} for visual comparisons.

\subsection{Limitation}
\begin{figure}
    \centering
    \begin{subfigure}{0.32\textwidth}
        \includegraphics[clip=true, trim={50 0 50 0}, width=\textwidth]{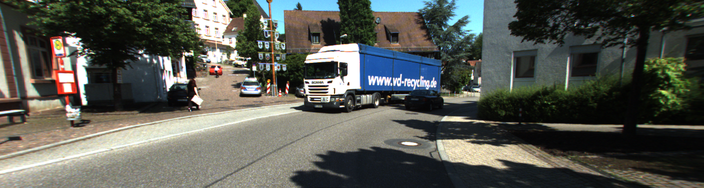}
        \caption*{GT}
    \end{subfigure}
    \begin{subfigure}{0.32\textwidth}
        \includegraphics[clip=true, trim={50 0 50 0}, width=\textwidth]{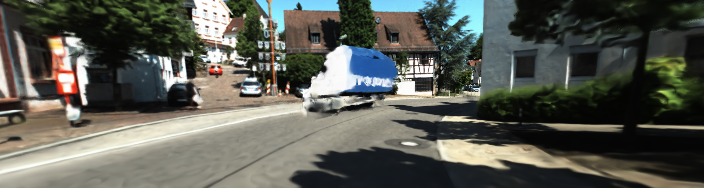}
        \caption*{$\hat{I}^S$}
    \end{subfigure}
    \begin{subfigure}{0.32\textwidth}
        \includegraphics[clip=true, trim={50 0 50 0}, width=\textwidth]{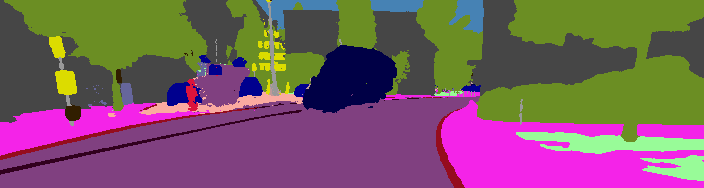}
        \caption*{$\hat{S}^S$}
    \end{subfigure}
    \begin{subfigure}{0.32\textwidth}
        \includegraphics[clip=true, trim={50 0 50 0}, width=\textwidth]{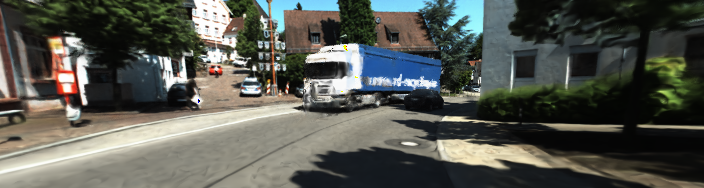}
        \caption*{$\hat{I}$}
    \end{subfigure}
    \begin{subfigure}{0.32\textwidth}
        \includegraphics[clip=true, trim={50 0 50 0}, width=\textwidth]{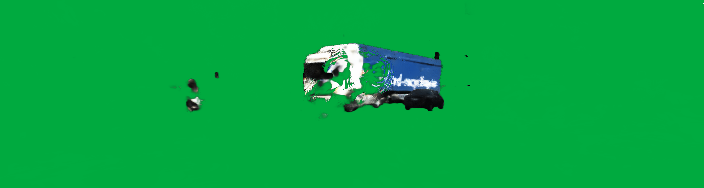}
        \caption*{$\hat{I}^D$}
    \end{subfigure}
    \begin{subfigure}{0.32\textwidth}
        \includegraphics[clip=true, trim={50 0 50 0}, width=\textwidth]{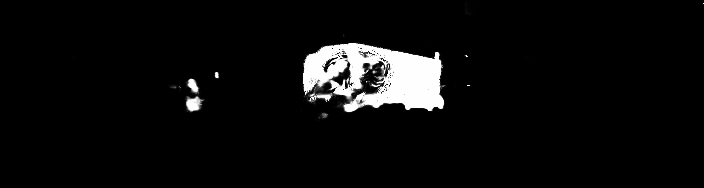}
        \caption*{$M^D$}
    \end{subfigure}
    \caption{\textbf{Failure case.} The static field is unable to recover regions that are highly occluded by the truck due to its lack of generalizing ability to unseen backgrounds. On the other hand, the dynamic field doesn't support object-level connection, leading to only half of the truck being visible in dynamic outputs.}
    \label{fig:limitation}
\end{figure}

In general, RoDUS works best with scenes where labels belonging to potentially movable objects are well-defined. Nonetheless, if the semantic ground truth is absent or heavily mislabeled, it moderately impacts performance.
Moreover, our approach can't handle areas that lack observations or have been highly occluded most of the time within the sequence. This is particularly notable by large or slow-moving objects in \cref{fig:limitation}. Exploring a model that is capable of uncertainty modeling to incorrect or anomalous pixels and inpainting these missing regions based on prior knowledge will be interesting as a future research direction.

\section{Conclusion}
In this paper, we presented RoDUS, a method for disentangling dynamic scenes using neural representation. Our approach effectively separated moving components from the scene without the need for manual labeling. To enhance disentangling efficiency, we proposed a robust training strategy and incorporated semantic reasoning, resulting in high-quality background view synthesis and accurate segmentation. Extensive experiments on challenging autonomous driving datasets showcased that RoDUS surpasses current state-of-the-art NeRF frameworks.

\bibliographystyle{splncs04}
\bibliography{main}

\newpage
\appendix

\section{Implementation Details}
\subsection{Data Processing}
\label{sec:data_process}

In our experiments, we downsample Pandaset images by a factor of 4 ($480 \times 270$) and KITTI-360 images by a factor of 2 ($704 \times 188$). The selected scenes from KITTI-360 are summarized in~\cref{tab:kitti_sequence}. For Pandaset, we use sequences 023, 042, and 043. We specifically select those scenes because they feature multiple dynamic objects while providing sufficient views of the background regions.

\begin{table}
    \caption{Selected KITTI-360 sequences for training, with skip step of 2.}
    \centering
    \setlength{\tabcolsep}{5pt}
    \begin{tabular}{ccccc}
    \toprule
        Sequence & Scene & Start Frame & End Frame & \#Frames per camera\\
    \midrule
        0 & 0002 & 4970 & 5230 & 130\\
        1 & 0003 & 100 & 330 & 115\\
        2 & 0007 & 2680 & 2930 & 125\\
        3 & 0010 & 910 & 1160 & 125\\
    \bottomrule
    \end{tabular}
    \label{tab:kitti_sequence}
\end{table}

For semantic awareness, the classes we assume to be dynamic and keep the contribution in the dynamic semantic head are: ``Person'', ``Rider'', ``Car'', ``Truck'', ``Bus'', ``Train'', ``Motorcycle'', ``Bicycle'', ``Bicyclist'', ``Motorcyclist'', and ``Unknown vehicle''.

\subsection{Architecture}
We adopt a proposal-based coarse-to-fine sampling technique\cite{barron2022mip} using two proposal models with the number of samples being $[128, 96]$ for the proposal models, and 48 for the final model. Our hash grids have the size of $2^{19}$ for $16$ levels with the base resolution and finest resolution of 16 and 2048 respectively. 
Additionally, our dynamic hash grid allows 4D inputs, which means that the intermediate feature vector is computed through a quadrilinear interpolation of 16 grid point vectors instead of 8 used in the 3D hash grid.
The static branch of our model is equipped with a 16-dimension per-frame embedding to adapt to varying lighting conditions.
 
\subsection{Losses}
The individual components used in the dynamic regularization loss are:

\noindent\textbf{Static-dynamic entropy factorization loss.}
We assume that any point in space can either belong to a static or dynamic object at one time, but not both, and encourage the transparency ratio at each point to be either 0 or 1, aiding the model in achieving a less entangled decomposition:
\begin{equation}
    \mathcal{L}_{\mathcal{H}} = \mathbb{E}\left[\mathcal{H}\left(\frac{\alpha^D}{\alpha^S+\alpha^D}\right)(\alpha^S+\alpha^D)\right],
\end{equation}

\noindent where $\mathcal{H}(x)=-(x\cdot\log(x)+(1-x)\cdot\log(1-x))$ is the binary entropy function and the term is weighted by the total transparency $(\alpha^S+\alpha^D)$ so that a point is allowed to have an empty density, as done in~\cite{yuan2021star}.

\noindent\textbf{Dynamic sparsity losses.}
The sparsity terms take as input densities $\sigma^D$ and shadow $\rho$ decoded from the dynamic field respectively and encourage those values to be sparse~\cite{wu2022d, sharma2023neural}:
\begin{equation}
    \mathcal{L}_{\sigma^D} = \mathbb{E}\left[\frac{1}{K}\sum_{i=1}^K\sigma_i^D\right]_{\mathbf{r}\in\mathcal{R}} ~~~\text{and}~~~ \mathcal{L}_{\rho} = \mathbb{E}\left[\frac{1}{K}\sum_{i=1}^KT_i\alpha_i\rho_i^2\right]_{\mathbf{r}\in\mathcal{R}}.
\end{equation}

\subsection{Training}

\begin{minipage}[t]{.49\textwidth}
We train RoDUS for 50 epochs using the Adam optimizer with a batch size of 200. The learning rate is set to $0.01$, and decayed to $0.0001$. The rays are grouped in $15\times 15$ patches to suit DSSIM, planar regularization, and robust losses.
\cref{tab:hyperparams} summarizes the hyperparameters used for the training of RoDUS, we do not require manual tuning for each scene, but we believe that such tuning could potentially further enhance performance.
\end{minipage}
\hfill
\begin{minipage}[t]{.49\textwidth}
    \captionof{table}{Training hyperparameters.}
    \centering
    \setlength{\tabcolsep}{5pt}
    \begin{tabular}{cc}
        \toprule
        Parameter & Value \\
        \midrule
        $\lambda_{d}$ & 0.5\\
        $\lambda_{sem}$ & 0.1\\
        $\lambda_{sky}$ & 0.03\\
        $\lambda_{road}$ & 0.1\\
        $\lambda_{\sigma^D}$ & 0.05\\
        $\lambda_{\rho}$ & 0.3\\
        $\lambda_{\mathcal{H}}$ & 0.01\\
        \bottomrule
    \end{tabular}
    \label{tab:hyperparams}
\end{minipage}
\vspace{0.5cm}

\noindent\textbf{Baseline Implementations.}
We adapt the code in the official repository of each method: RobustNeRF\footnote{\url{https://robustnerf.github.io/}}, D$^2$NeRF\footnote{\url{https://github.com/ChikaYan/d2nerf}}, SUDS\footnote{\url{https://github.com/hturki/suds}}, EmerNeRF\footnote{\url{https://github.com/NVlabs/EmerNeRF}} and modify their dataloaders. We search for the best robust loss threshold $\mathcal{T}_\epsilon$ for RobustNeRF, while the rest are trained with the default parameters provided in the repositories.

\section{Additional Results}
\begin{figure*}[tp]
\centering
\scriptsize
\setlength{\tabcolsep}{0.002\linewidth}
\renewcommand{\arraystretch}{0.8}
\begin{tabular}{ccccc} 
    \multirow{1}{*}[10mm]{\rotatebox[origin=c]{90}{\scriptsize GT}} & &
    \includegraphics[width=0.30\textwidth]{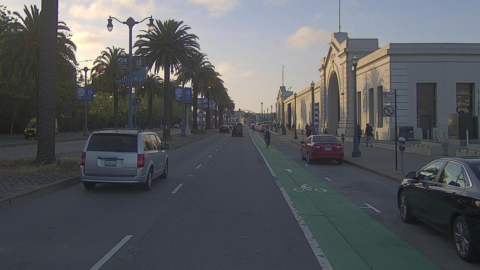} &  & \\

    \multirow{1}{*}[16mm]{\rotatebox[origin=c]{90}{\scriptsize RobustNeRF}} &
    \multirow{1}{*}[10mm]{\rotatebox[origin=c]{90}{\scriptsize \cite{sabour2023robustnerf}}} &
    \framebox(0.29\linewidth,0.165\linewidth){\small N/A} & 
    \includegraphics[width=0.30\textwidth]{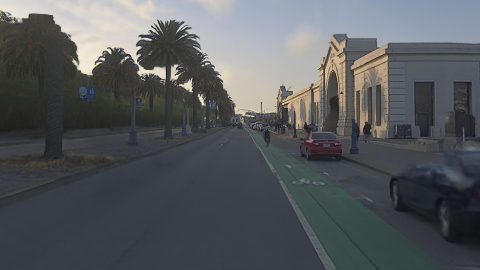} & 
    \framebox(0.29\linewidth,0.165\linewidth){\small N/A} \\

    \multirow{1}{*}[12mm]{\rotatebox[origin=c]{90}{\scriptsize D$^2$NeRF}} &
    \multirow{1}{*}[10mm]{\rotatebox[origin=c]{90}{\scriptsize \cite{wu2022d}}} &
    \includegraphics[width=0.30\textwidth]{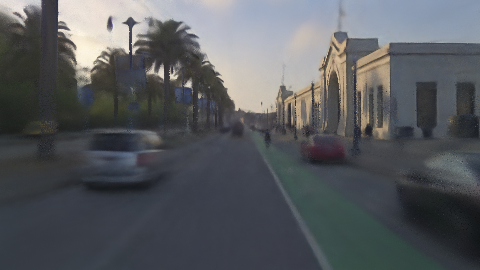} & 
    \includegraphics[width=0.30\textwidth]{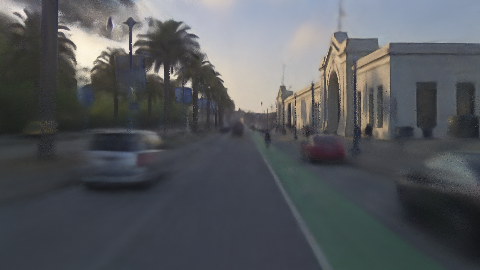} & 
    \includegraphics[width=0.30\textwidth]{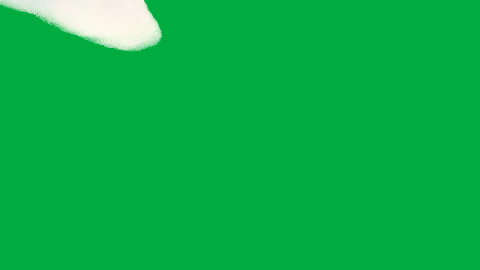} \\

    \multirow{1}{*}[12mm]{\rotatebox[origin=c]{90}{\scriptsize SUDS}} &
    \multirow{1}{*}[10mm]{\rotatebox[origin=c]{90}{\scriptsize \cite{turki2023suds}}} &
    \includegraphics[width=0.30\textwidth]{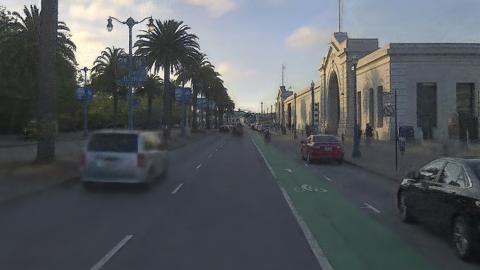} & 
    \includegraphics[width=0.30\textwidth]{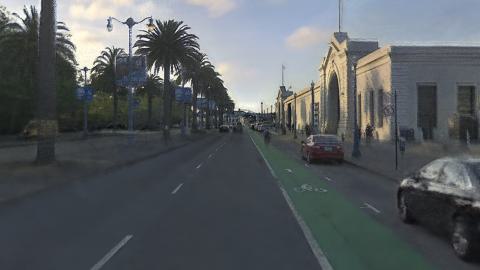} & 
    \includegraphics[width=0.30\textwidth]{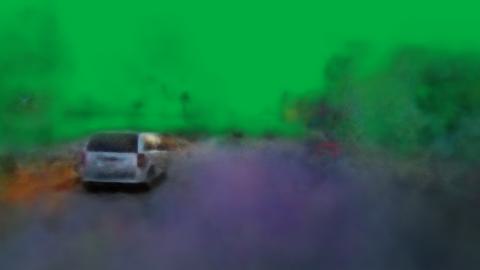} \\

    \multirow{1}{*}[14mm]{\rotatebox[origin=c]{90}{\scriptsize EmerNeRF}} &
    \multirow{1}{*}[10mm]{\rotatebox[origin=c]{90}{\scriptsize \cite{yang2023emernerf}}} &
    \includegraphics[width=0.30\textwidth]{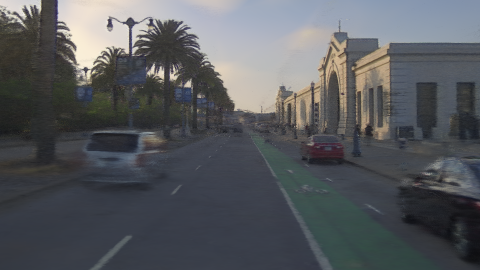} & 
    \includegraphics[width=0.30\textwidth]{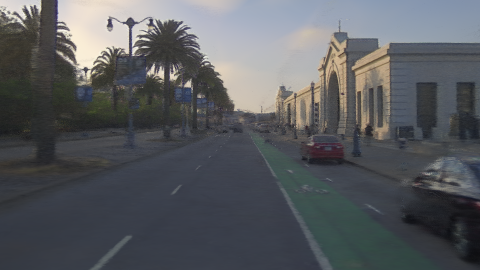} & 
    \includegraphics[width=0.30\textwidth]{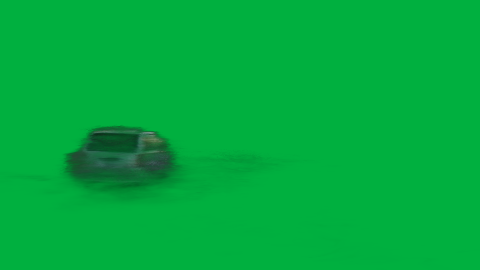} \\

    \multirow{1}{*}[12mm]{\rotatebox[origin=c]{90}{\scriptsize RoDUS}} &
    \multirow{1}{*}[12mm]{\rotatebox[origin=c]{90}{\scriptsize (ours)}} &
    \includegraphics[width=0.30\textwidth]{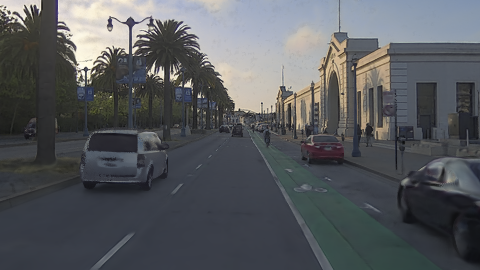} & 
    \includegraphics[width=0.30\textwidth]{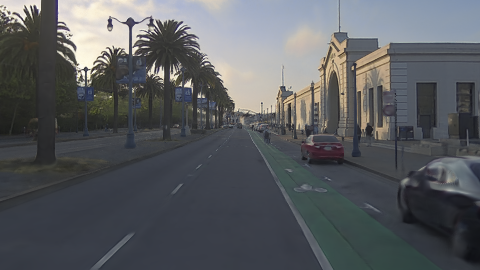} & 
    \includegraphics[width=0.30\textwidth]{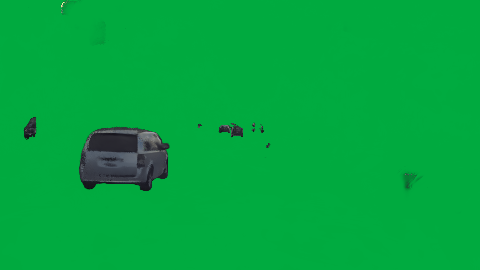} \\           
    & ~ & \small Composed RGB & \small Static RGB & \small Dynamic RGB \\
            
\end{tabular}
\caption{Additional qualitative comparisons on Pandaset.}
\label{fig:more_panda_results}
\end{figure*}

\begin{figure}[tp]
\setlength{\tabcolsep}{0.002\linewidth}
\renewcommand{\arraystretch}{0.8}
\begin{tabular}{ccccc}
    
    \multirow{1}{*}[5.7mm]{\rotatebox[origin=c]{90}{\scriptsize GT}} & &
    \includegraphics[clip=true, trim={50 0 50 0},width=0.32\textwidth]{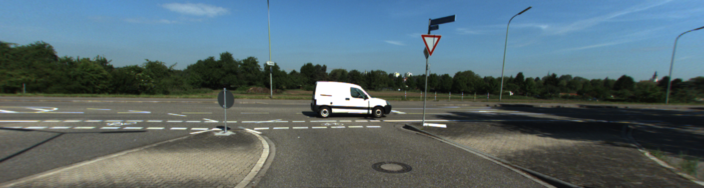} & &
    \\

    \multirow{1}{*}[12mm]{\rotatebox[origin=c]{90}{\scriptsize RobustNeRF}} &
    \multirow{1}{*}[6mm]{\rotatebox[origin=c]{90}{\scriptsize \cite{sabour2023robustnerf}}} &
    \framebox(0.32\linewidth,0.098\linewidth){N/A} & 
    \includegraphics[clip=true, trim={50 0 50 0},width=0.32\textwidth]{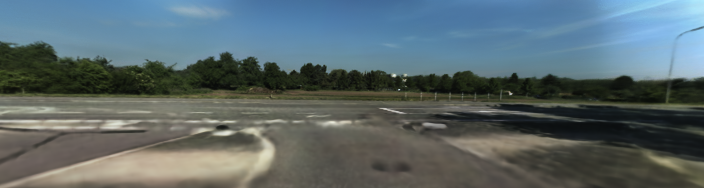} & 
    \framebox(0.32\linewidth,0.098\linewidth){N/A}
    \\

    \multirow{1}{*}[7.5mm]{\rotatebox[origin=c]{90}{\scriptsize D$^2$NeRF}} &
    \multirow{1}{*}[6mm]{\rotatebox[origin=c]{90}{\scriptsize \cite{wu2022d}}} &
    \includegraphics[clip=true, trim={50 0 50 0},width=0.32\textwidth]{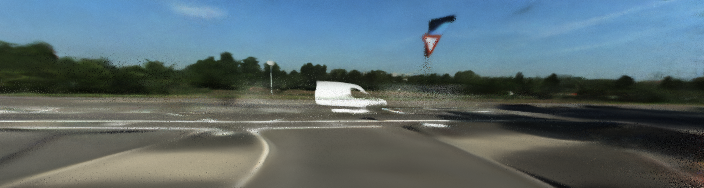} & 
    \includegraphics[clip=true, trim={50 0 50 0},width=0.32\textwidth]{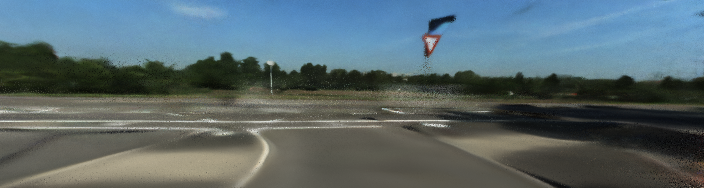} & 
    \includegraphics[clip=true, trim={50 0 50 0},width=0.32\textwidth]{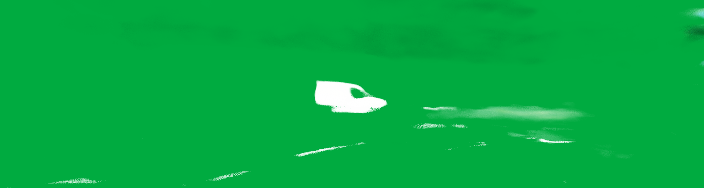}
    \\

    \multirow{1}{*}[7.5mm]{\rotatebox[origin=c]{90}{\scriptsize SUDS}} &
    \multirow{1}{*}[6mm]{\rotatebox[origin=c]{90}{\scriptsize \cite{turki2023suds}}} &
    \includegraphics[clip=true, trim={50 0 50 0},width=0.32\textwidth]{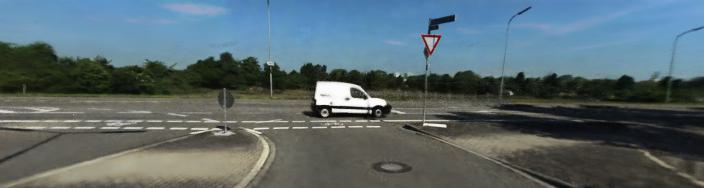} & 
    \includegraphics[clip=true, trim={50 0 50 0},width=0.32\textwidth]{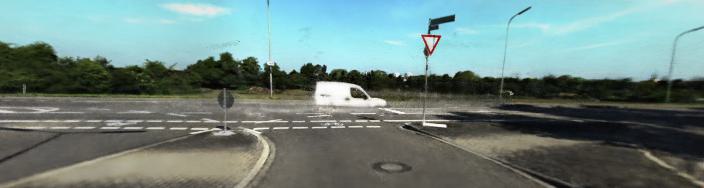} & 
    \includegraphics[clip=true, trim={50 0 50 0},width=0.32\textwidth]{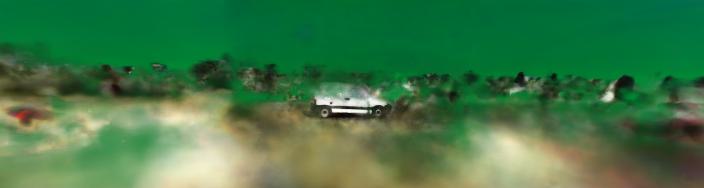}
    \\

    \multirow{1}{*}[10mm]{\rotatebox[origin=c]{90}{\scriptsize EmerNeRF}} &
    \multirow{1}{*}[6mm]{\rotatebox[origin=c]{90}{\scriptsize \cite{yang2023emernerf}}} &
    \includegraphics[clip=true, trim={50 0 50 0},width=0.32\textwidth]{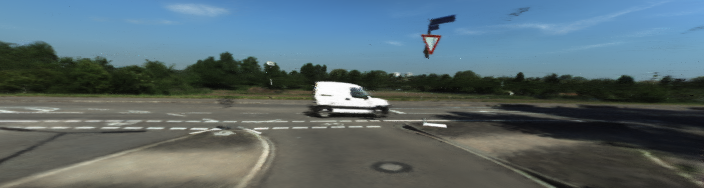} & 
    \includegraphics[clip=true, trim={50 0 50 0},width=0.32\textwidth]{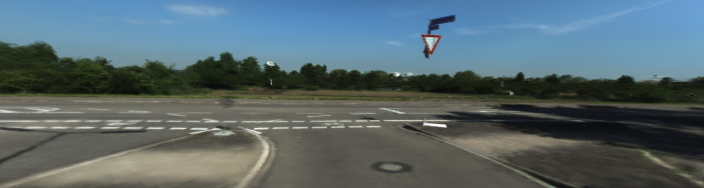} & 
    \includegraphics[clip=true, trim={50 0 50 0},width=0.32\textwidth]{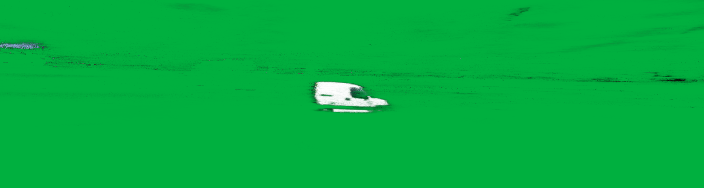}
    \\

    \multirow{1}{*}[8mm]{\rotatebox[origin=c]{90}{\scriptsize RoDUS}} &
    \multirow{1}{*}[7mm]{\rotatebox[origin=c]{90}{\scriptsize (ours)}} &
    \includegraphics[clip=true, trim={50 0 50 0},width=0.32\textwidth]{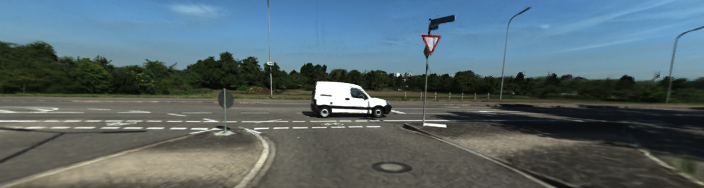} & 
    \includegraphics[clip=true, trim={50 0 50 0},width=0.32\textwidth]{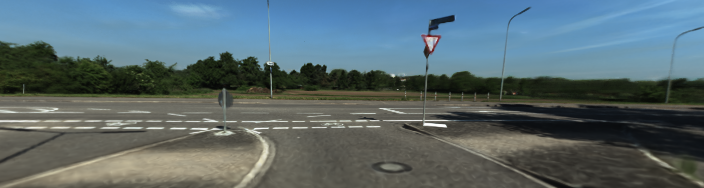} & 
    \includegraphics[clip=true, trim={50 0 50 0},width=0.32\textwidth]{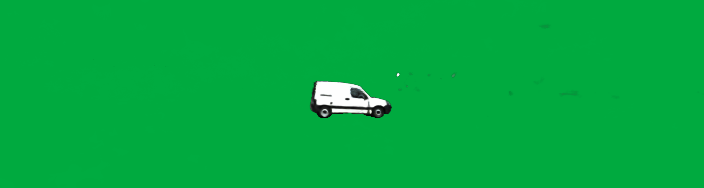}
    \\

    & & & \\
    \hline
    & & & \\
    
    \multirow{1}{*}[5.7mm]{\rotatebox[origin=c]{90}{\scriptsize GT}} &
    \multirow{1}{*}[6.5mm]{\rotatebox[origin=c]{90}{}} &
    \includegraphics[clip=true, trim={50 0 50 0},width=0.32\textwidth]{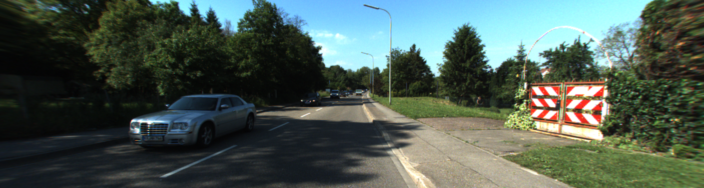} & &
    \\

    \multirow{1}{*}[12mm]{\rotatebox[origin=c]{90}{\scriptsize RobustNeRF}} &
    \multirow{1}{*}[6mm]{\rotatebox[origin=c]{90}{\scriptsize \cite{sabour2023robustnerf}}} &
    \framebox(0.32\linewidth,0.098\linewidth){N/A} & 
    \includegraphics[clip=true, trim={50 0 50 0},width=0.32\textwidth]{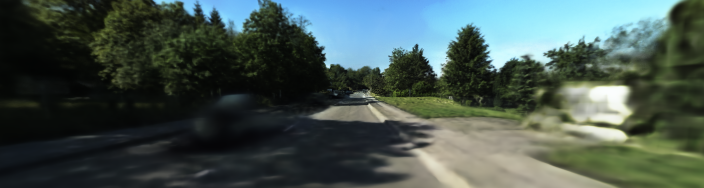} & 
    \framebox(0.32\linewidth,0.098\linewidth){N/A}
    \\

    \multirow{1}{*}[7.5mm]{\rotatebox[origin=c]{90}{\scriptsize D$^2$NeRF}} &
    \multirow{1}{*}[6mm]{\rotatebox[origin=c]{90}{\scriptsize \cite{wu2022d}}} &
    \includegraphics[clip=true, trim={50 0 50 0},width=0.32\textwidth]{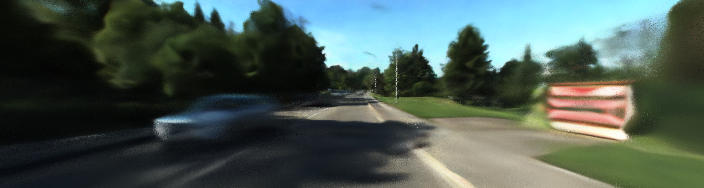} & 
    \includegraphics[clip=true, trim={50 0 50 0},width=0.32\textwidth]{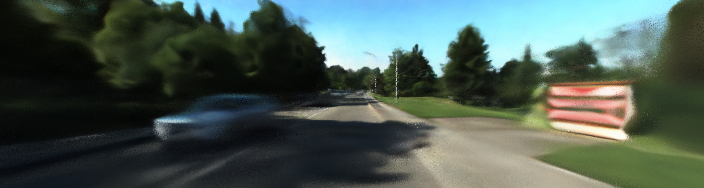} & 
    \includegraphics[clip=true, trim={50 0 50 0},width=0.32\textwidth]{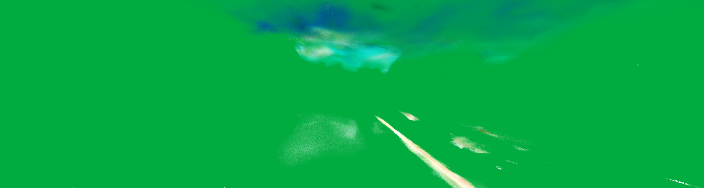}
    \\

    \multirow{1}{*}[7.5mm]{\rotatebox[origin=c]{90}{\scriptsize SUDS}} &
    \multirow{1}{*}[6mm]{\rotatebox[origin=c]{90}{\scriptsize \cite{turki2023suds}}} &
    \includegraphics[clip=true, trim={50 0 50 0},width=0.32\textwidth]{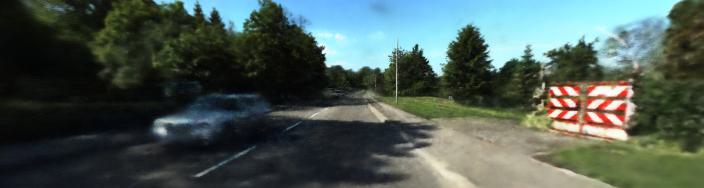} & 
    \includegraphics[clip=true, trim={50 0 50 0},width=0.32\textwidth]{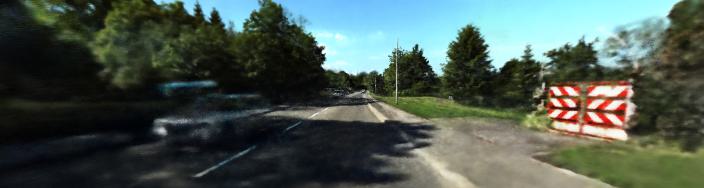} & 
    \includegraphics[clip=true, trim={50 0 50 0},width=0.32\textwidth]{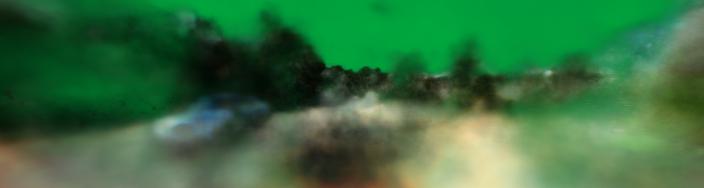}
    \\

    \multirow{1}{*}[10mm]{\rotatebox[origin=c]{90}{\scriptsize EmerNeRF}} &
    \multirow{1}{*}[6mm]{\rotatebox[origin=c]{90}{\scriptsize \cite{yang2023emernerf}}} &
    \includegraphics[clip=true, trim={50 0 50 0},width=0.32\textwidth]{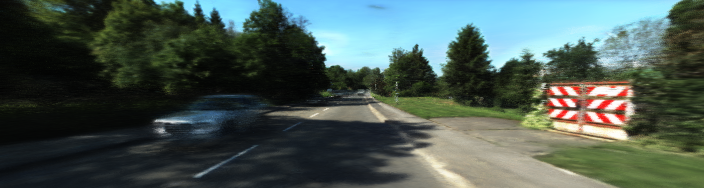} & 
    \includegraphics[clip=true, trim={50 0 50 0},width=0.32\textwidth]{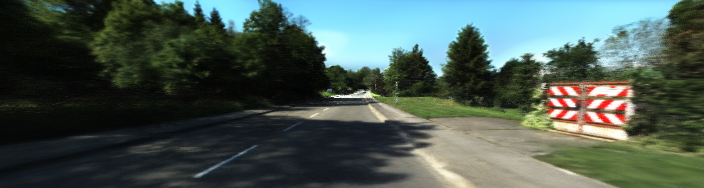} & 
    \includegraphics[clip=true, trim={50 0 50 0},width=0.32\textwidth]{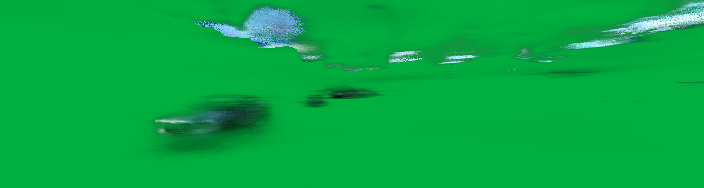}
    \\

    \multirow{1}{*}[8mm]{\rotatebox[origin=c]{90}{\scriptsize RoDUS}} &
    \multirow{1}{*}[7mm]{\rotatebox[origin=c]{90}{\scriptsize (ours)}} &
    \includegraphics[clip=true, trim={50 0 50 0},width=0.32\textwidth]{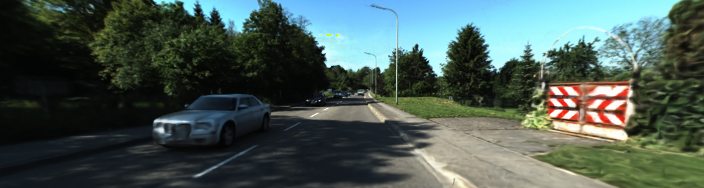} & 
    \includegraphics[clip=true, trim={50 0 50 0},width=0.32\textwidth]{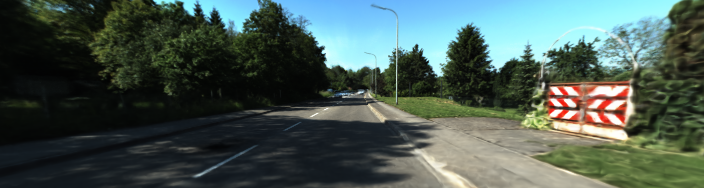} & 
    \includegraphics[clip=true, trim={50 0 50 0},width=0.32\textwidth]{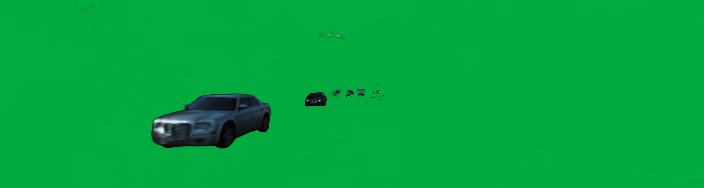}
    \\
        
    & ~ & Compose RGB & Static RGB & Dynamic RGB \\
    
    \end{tabular}
\caption{Additional qualitative comparisons on KITTI-360.}
\label{fig:more_kitti_results}
\end{figure}

\begin{figure}[tp]
\begin{minipage}[b]{0.03\textwidth}
     \rotatebox{90}{GT}
\end{minipage}
\hfill
\raisebox{-1\baselineskip}{
\begin{subfigure}[b]{0.32\linewidth}
    \includegraphics[clip=true, trim={50 0 50 0}, width=\linewidth]{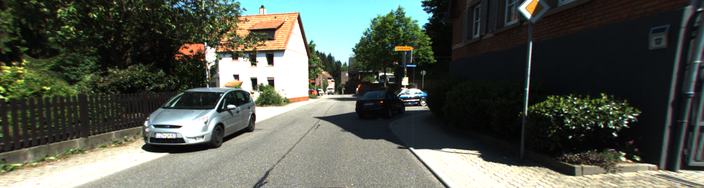}
\end{subfigure}
\begin{subfigure}[b]{0.32\linewidth}
    \includegraphics[clip=true, trim={50 0 50 0}, width=\linewidth]{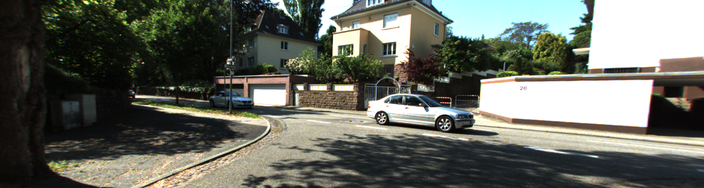}
\end{subfigure}
\begin{subfigure}[b]{0.32\linewidth}
    \includegraphics[clip=true, trim={50 0 50 0}, width=\linewidth]{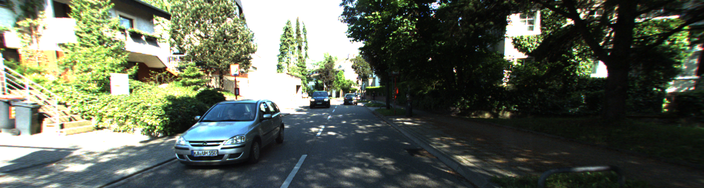}
\end{subfigure}}

\begin{minipage}[b]{0.03\textwidth}
     \rotatebox{90}{$\hat{I}$}
\end{minipage}
\hfill
\raisebox{-1\baselineskip}{
\begin{subfigure}[b]{0.32\linewidth}
    \includegraphics[clip=true, trim={50 0 50 0}, width=\linewidth]{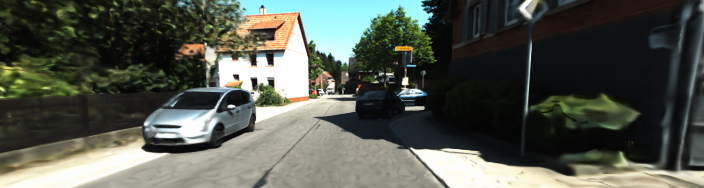}
\end{subfigure}
\begin{subfigure}[b]{0.32\linewidth}
    \includegraphics[clip=true, trim={50 0 50 0}, width=\linewidth]{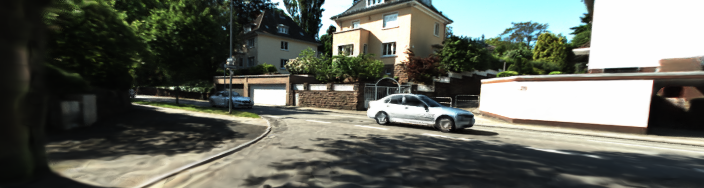}
\end{subfigure}
\begin{subfigure}[b]{0.32\linewidth}
    \includegraphics[clip=true, trim={50 0 50 0}, width=\linewidth]{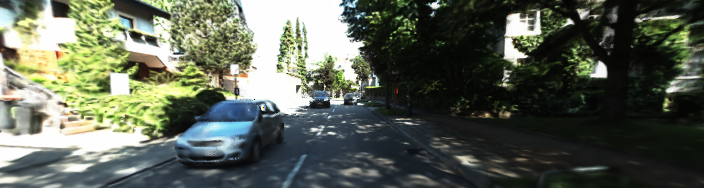}
\end{subfigure}}

\begin{minipage}[b]{0.03\textwidth}
     \rotatebox{90}{$\hat{d}$}
\end{minipage}
\hfill
\raisebox{-1\baselineskip}{
\begin{subfigure}[b]{0.32\linewidth}
    \includegraphics[clip=true, trim={50 0 50 0}, width=\linewidth]{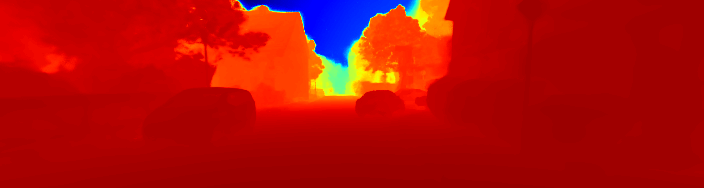}
\end{subfigure}
\begin{subfigure}[b]{0.32\linewidth}
    \includegraphics[clip=true, trim={50 0 50 0}, width=\linewidth]{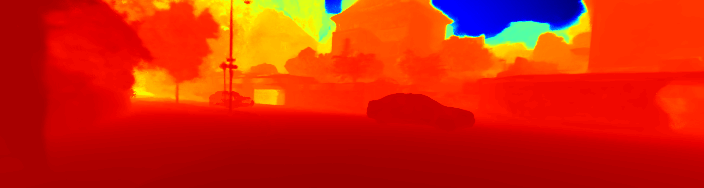}
\end{subfigure}
\begin{subfigure}[b]{0.32\linewidth}
    \includegraphics[clip=true, trim={50 0 50 0}, width=\linewidth]{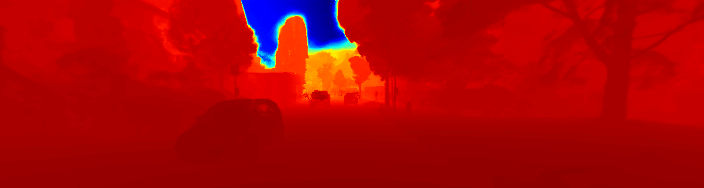}
\end{subfigure}}

\begin{minipage}[b]{0.03\textwidth}
     \rotatebox{90}{$\hat{I}^S$}
\end{minipage}
\hfill
\raisebox{-1\baselineskip}{
\begin{subfigure}[b]{0.32\linewidth}
    \includegraphics[clip=true, trim={50 0 50 0}, width=\linewidth]{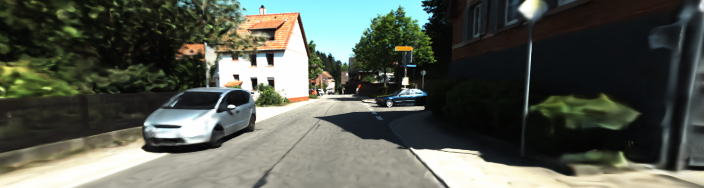}
\end{subfigure}
\begin{subfigure}[b]{0.32\linewidth}
    \includegraphics[clip=true, trim={50 0 50 0}, width=\linewidth]{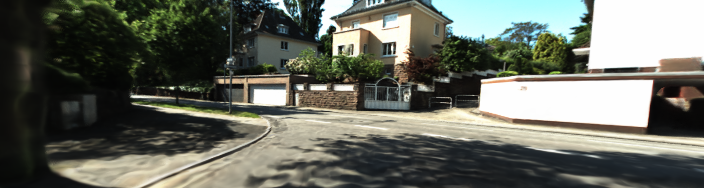}
\end{subfigure}
\begin{subfigure}[b]{0.32\linewidth}
    \includegraphics[clip=true, trim={50 0 50 0}, width=\linewidth]{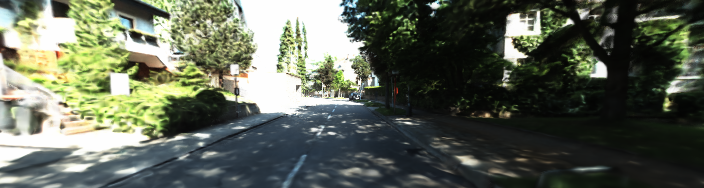}
\end{subfigure}}

\begin{minipage}[b]{0.03\textwidth}
     \rotatebox{90}{$\hat{d}^S$}
\end{minipage}
\hfill
\raisebox{-1\baselineskip}{
\begin{subfigure}[b]{0.32\linewidth}
    \includegraphics[clip=true, trim={50 0 50 0}, width=\linewidth]{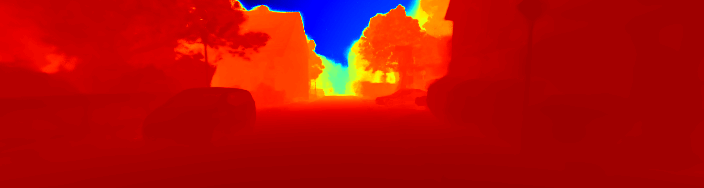}
\end{subfigure}
\begin{subfigure}[b]{0.32\linewidth}
    \includegraphics[clip=true, trim={50 0 50 0}, width=\linewidth]{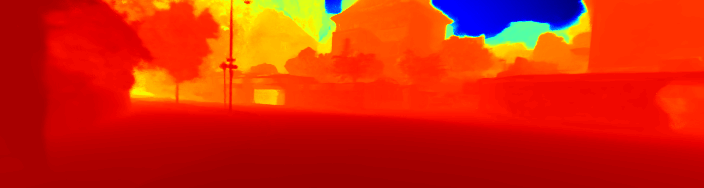}
\end{subfigure}
\begin{subfigure}[b]{0.32\linewidth}
    \includegraphics[clip=true, trim={50 0 50 0}, width=\linewidth]{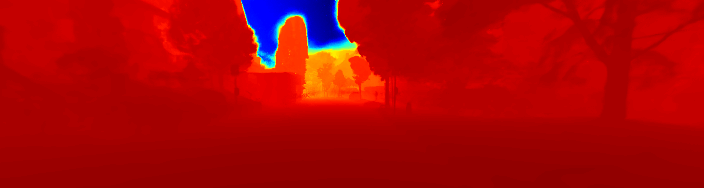}
\end{subfigure}}

\begin{minipage}[b]{0.03\textwidth}
     \rotatebox{90}{$\hat{I}^D$}
\end{minipage}
\hfill
\raisebox{-1\baselineskip}{
\begin{subfigure}[b]{0.32\linewidth}
    \includegraphics[clip=true, trim={50 0 50 0}, width=\linewidth]{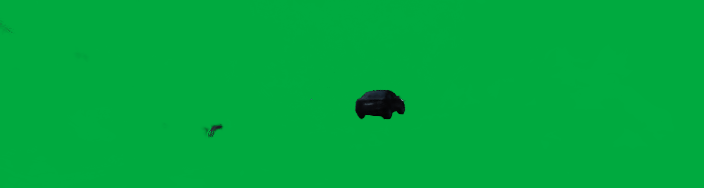}
\end{subfigure}
\begin{subfigure}[b]{0.32\linewidth}
    \includegraphics[clip=true, trim={50 0 50 0}, width=\linewidth]{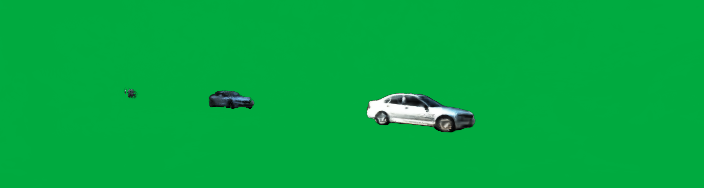}
\end{subfigure}
\begin{subfigure}[b]{0.32\linewidth}
    \includegraphics[clip=true, trim={50 0 50 0}, width=\linewidth]{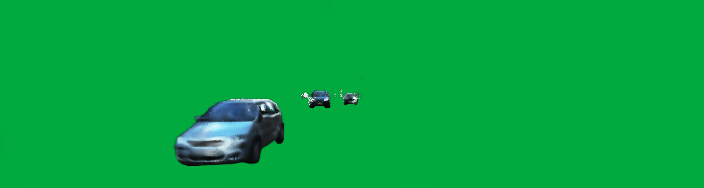}
\end{subfigure}}

\begin{minipage}[b]{0.03\textwidth}
     \rotatebox{90}{$\hat{S}$}
\end{minipage}
\hfill
\raisebox{-1\baselineskip}{
\begin{subfigure}[b]{0.32\linewidth}
    \includegraphics[clip=true, trim={50 0 50 0}, width=\linewidth]{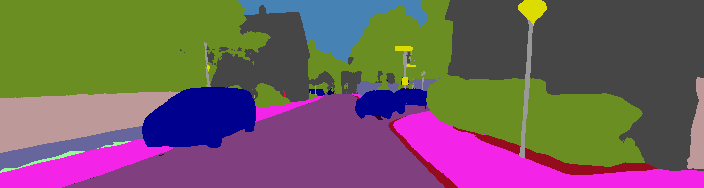}
\end{subfigure}
\begin{subfigure}[b]{0.32\linewidth}
    \includegraphics[clip=true, trim={50 0 50 0}, width=\linewidth]{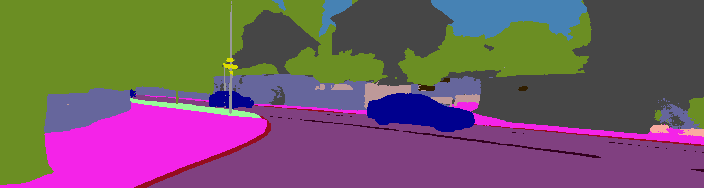}
\end{subfigure}
\begin{subfigure}[b]{0.32\linewidth}
    \includegraphics[clip=true, trim={50 0 50 0}, width=\linewidth]{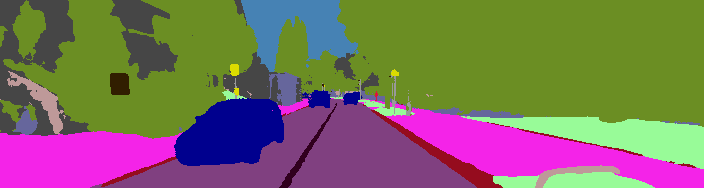}
\end{subfigure}}

\begin{minipage}[b]{0.03\textwidth}
     \rotatebox{90}{$\hat{S}^S$}
\end{minipage}
\hfill
\raisebox{-1\baselineskip}{
\begin{subfigure}[b]{0.32\linewidth}
    \includegraphics[clip=true, trim={50 0 50 0}, width=\linewidth]{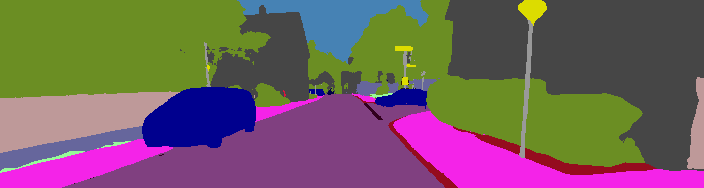}
\end{subfigure}
\begin{subfigure}[b]{0.32\linewidth}
    \includegraphics[clip=true, trim={50 0 50 0}, width=\linewidth]{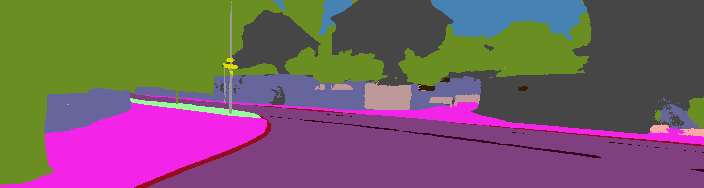}
\end{subfigure}
\begin{subfigure}[b]{0.32\linewidth}
    \includegraphics[clip=true, trim={50 0 50 0}, width=\linewidth]{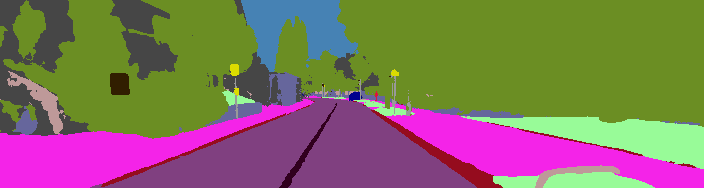}
\end{subfigure}}

\begin{minipage}[b]{0.03\textwidth}
     \rotatebox{90}{$M^D$}
\end{minipage}
\hfill
\raisebox{-1\baselineskip}{
\begin{subfigure}[b]{0.32\linewidth}
    \includegraphics[clip=true, trim={50 0 50 0}, width=\linewidth]{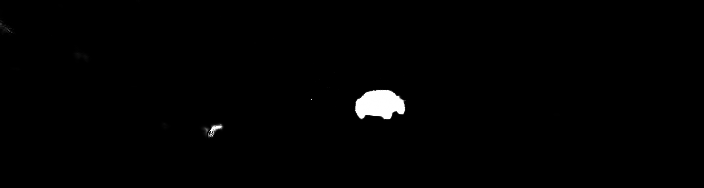}
\end{subfigure}
\begin{subfigure}[b]{0.32\linewidth}
    \includegraphics[clip=true, trim={50 0 50 0}, width=\linewidth]{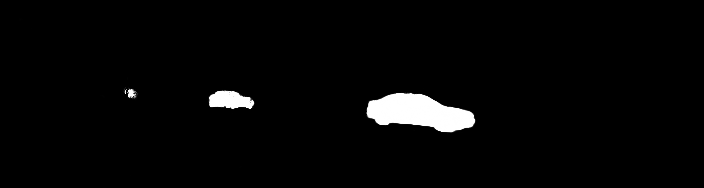}
\end{subfigure}
\begin{subfigure}[b]{0.32\linewidth}
    \includegraphics[clip=true, trim={50 0 50 0}, width=\linewidth]{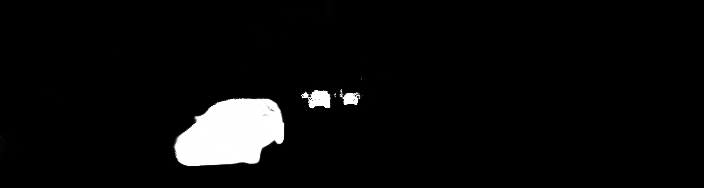}
\end{subfigure}}
    \caption{Additional showcases of RoDUS's capability.}
    \label{fig:more_results}
\end{figure}

\cref{fig:more_panda_results} and \cref{fig:more_kitti_results} show additional qualitative comparisons between our
RoDUS and previous methods in static-dynamic decomposition and scene reconstruction. \cref{fig:more_results} displays RoDUS's performance on other different scenes, demonstrating that RoDUS's dynamic branch is able to classify moving objects from parked vehicles or other static instances. Furthermore, RoDUS can be trained from noisy semantic annotations, leading to a scene-consistent semantic representation. Our combination of multiple modalities also supports the model in reconstructing and segmenting thin structures and objects (\eg light poles, and traffic signs).

\section{Additional Ablation and Analysis}
\subsection{Robust Kernel Threshold}
\begin{figure}
    \centering
    \begin{subfigure}[b]{0.30\linewidth}
        \includegraphics[width=\linewidth]{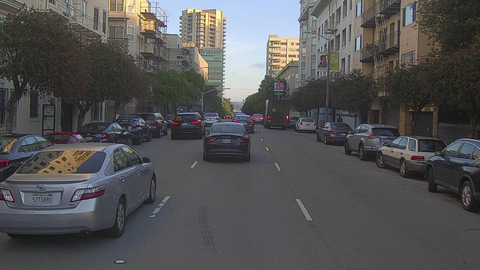}
        \caption*{GT}
    \end{subfigure}
    \begin{subfigure}[b]{0.30\linewidth}
        \includegraphics[width=\linewidth]{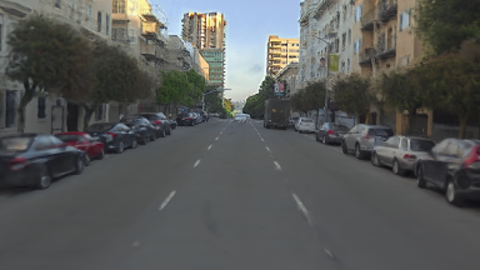}
        \caption*{RoDUS's static RGB}
    \end{subfigure}\\
    
    \begin{subfigure}[b]{0.30\linewidth}
        \includegraphics[width=\linewidth]{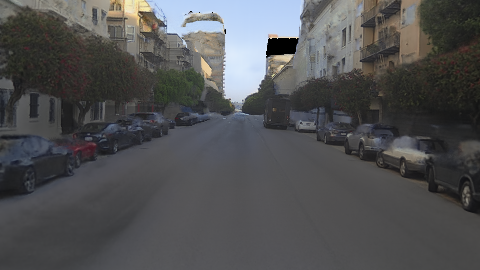}
        \caption*{RobustNeRF ($\mathcal{T}_\epsilon=0.7$)}
    \end{subfigure}
    \begin{subfigure}[b]{0.30\linewidth}
        \includegraphics[width=\linewidth]{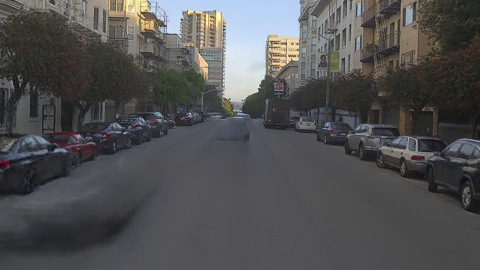}
        \caption*{RobustNeRF ($\mathcal{T}_\epsilon=0.9$)}
    \end{subfigure}
    \begin{subfigure}[b]{0.30\linewidth}
        \includegraphics[width=\linewidth]{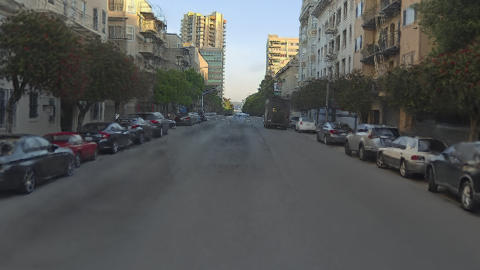}
        \caption*{RobustNeRF ($\mathcal{T}_\epsilon=0.75$)}
    \end{subfigure}
    \caption{Qualitative comparison on static RGB image with different threshold values.}
    \label{fig:threshold}
\end{figure}

We explain the choice of percentile threshold $\mathcal{T}_\epsilon$ used in RobustNeRF~\cite{sabour2023robustnerf} and report qualitative results in \cref{fig:threshold}. 
With $\mathcal{T}_\epsilon=0.7$, there are still a lot of structures that could not be learned (\eg the far away building), when we increase $\mathcal{T}_\epsilon$ to $0.9$, the shadow regions start to appear. As a result, we select $\mathcal{T}_\epsilon=0.75$ for benchmarking as it yields the best visual quality with the least artifacts. Correspondingly, the kernel saturates and we observed no improvement as the training continues. Notice that the model still could not learn the road marking in any case. 

\subsection{Various Challenging Scenarios}
\begin{figure*}[tp]
\centering
\scriptsize
\setlength{\tabcolsep}{0.002\linewidth}
\renewcommand{\arraystretch}{0.8}
\begin{tabular}{ccccc} 
    \multirow{1}{*}[18mm]{\rotatebox[origin=c]{90}{\scriptsize Complex traffic}} &
    \multirow{1}{*}[10mm]{\rotatebox[origin=c]{90}{}} &
    \includegraphics[width=0.30\textwidth]{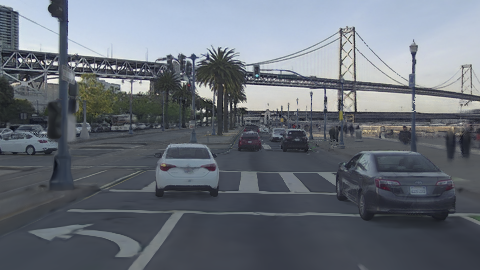} & 
    \includegraphics[width=0.30\textwidth]{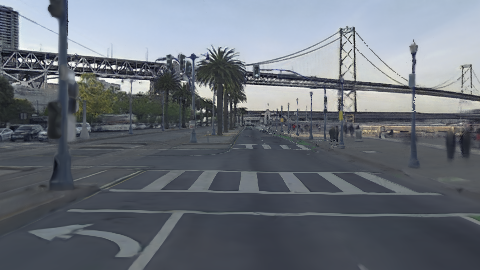} & 
    \includegraphics[width=0.30\textwidth]{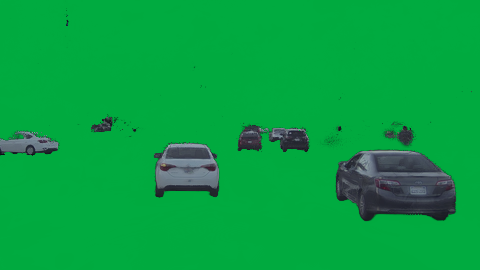} \\

    \multirow{1}{*}[12mm]{\rotatebox[origin=c]{90}{\scriptsize Cyclist}} &
    \multirow{1}{*}[10mm]{\rotatebox[origin=c]{90}{}} &
    \includegraphics[width=0.30\textwidth]{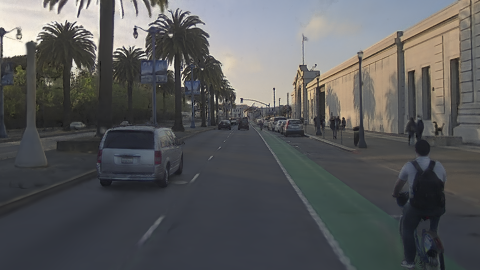} & 
    \includegraphics[width=0.30\textwidth]{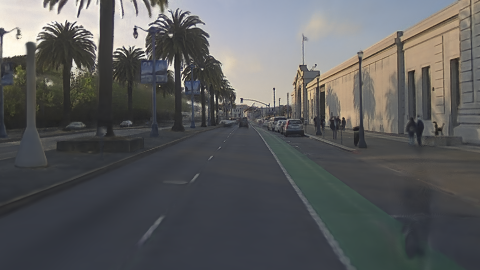} & 
    \includegraphics[width=0.30\textwidth]{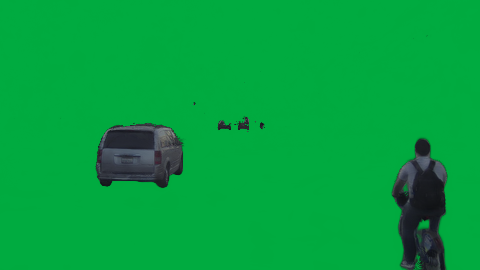} \\
    
    \multirow{1}{*}[14mm]{\rotatebox[origin=c]{90}{\scriptsize Low light}} &
    \multirow{1}{*}[10mm]{\rotatebox[origin=c]{90}{}} &
    \includegraphics[width=0.30\textwidth]{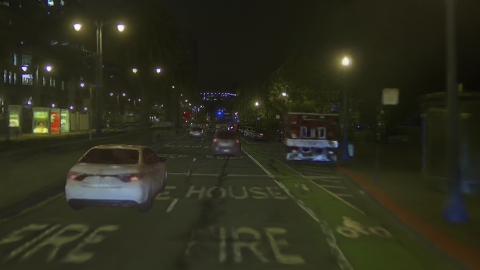} & 
    \includegraphics[width=0.30\textwidth]{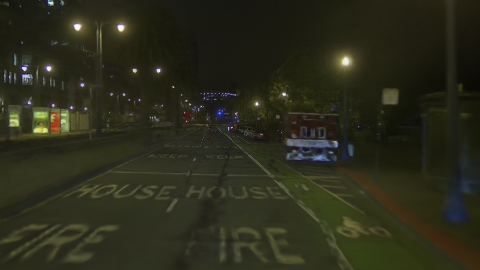} & 
    \includegraphics[width=0.30\textwidth]{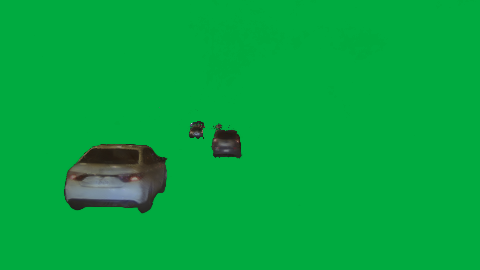} \\

    \multirow{1}{*}[10mm]{\rotatebox[origin=c]{90}{\scriptsize Uneven}} &
    \multirow{1}{*}[12mm]{\rotatebox[origin=c]{90}{\scriptsize Illumination}} &
    \includegraphics[clip=true, trim={100 0 100 0},width=0.30\textwidth]{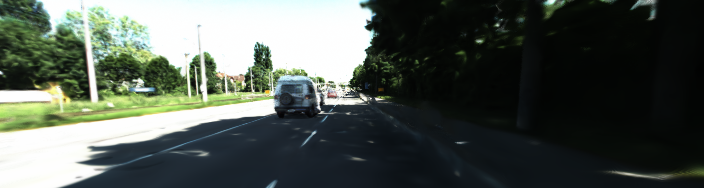} & 
    \includegraphics[clip=true, trim={100 0 100 0},width=0.30\textwidth]{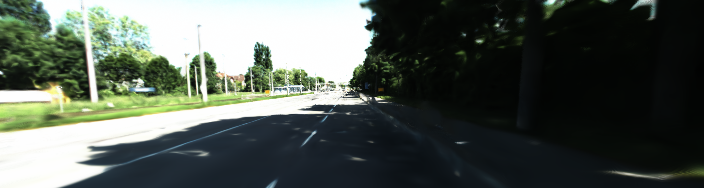} & 
    \includegraphics[clip=true, trim={100 0 100 0},width=0.30\textwidth]{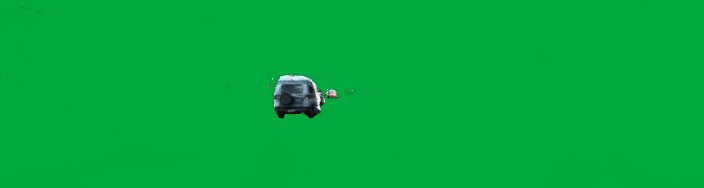} \\
 
    & ~ & \small Composed RGB & \small Static RGB & \small Dynamic RGB \\
            
\end{tabular}
\caption{RoDUS’s adaptivity on different scenarios.}
\label{fig:conditions}
\end{figure*}

\cref{fig:conditions} illustrates RoDUS's performance in low light (driving at night) and uneven illumination (driving into a shaded area) environments with high variations between views which are common in an autonomous driving context. It also highlights that the model could handle chaotic traffic and slow-moving entities (\ie cyclists) as long as there is no significant occlusion.

\subsection{Shadow Handling}
\begin{figure*}[tp]
    \centering
    \scriptsize
    \setlength{\tabcolsep}{0.002\linewidth}
    \renewcommand{\arraystretch}{0.8}
    \begin{tabular}{ccccc}
        \multirow{1}{*}[8mm]{\rotatebox[origin=c]{90}{\scriptsize GT}} & &
        \includegraphics[clip=true, trim={100 0 0 0},width=0.32\textwidth]{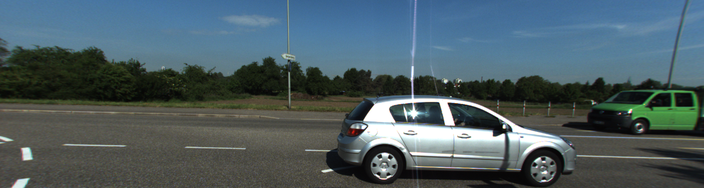} & 
        \includegraphics[clip=true, trim={50 0 50 0},width=0.32\textwidth]{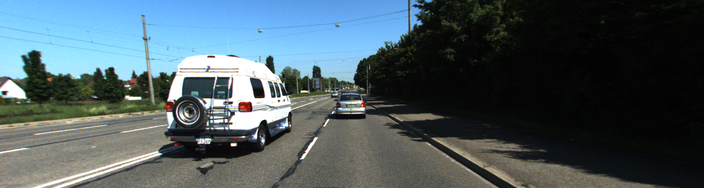} & 
        \includegraphics[clip=true, trim={0 0 100 0},width=0.32\textwidth]{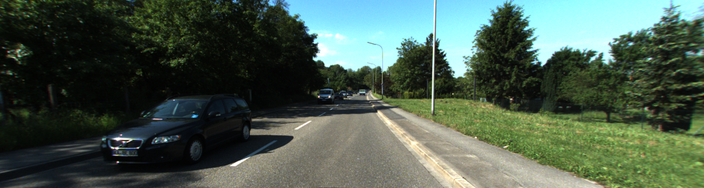}
        \\

        \multirow{1}{*}[10mm]{\rotatebox[origin=c]{90}{\scriptsize Shadow}} &
        \multirow{1}{*}[8mm]{\rotatebox[origin=c]{90}{\scriptsize mask}} &
        \includegraphics[clip=true, trim={100 0 0 0},width=0.32\textwidth]{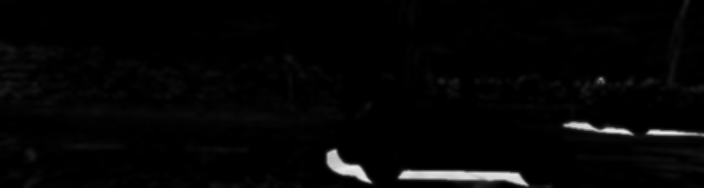} & 
        \includegraphics[clip=true, trim={50 0 50 0},width=0.32\textwidth]{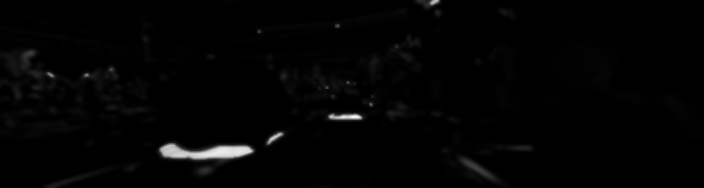} & 
        \includegraphics[clip=true, trim={0 0 100 0},width=0.32\textwidth]{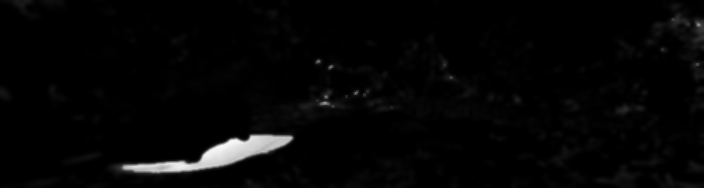}
        \\
    
    \end{tabular}
    \caption{We obtain the shadow mask by applying volumetric rendering on the shadow ratio $\rho$, demonstrating that the model is learning the correct shadow regions.}
    \label{fig:shadow}
\end{figure*}

Shadows cast by dynamic objects are also time-varying and modeled separately in our model. \cref{fig:shadow} shows that RoDUS successfully decouples the shadows from the scene.

\section{Discussions}
\subsection{Ethical Concern}
Our method mainly serves the purpose of removing dynamic vehicles from the scenes. While such methods hold promise in augmenting training environments for autonomous vehicles by simulating diverse scenarios, considering the act of modifying real-world driving data for inappropriate motives (\eg erasing actors, generating fake content) is \textbf{not} encouraged. Altering records for any purpose carries the risk of distorting the true nature of driving events and the credibility of data-driven insights.
In contrast, we would be delighted to see directions that focus on detecting image forgery for such methods. We believe this area of study could hold significant potential for advancements in the field, providing opportunities to enhance the robustness and security of digital content.

\subsection{Limitations}
Apart from the main limitation mentioned in the paper, we also encountered several problems that are related to most NeRF-based methods in general: RoDUS demands precise camera poses, which require multiple sources of odometry when working with outdoor, high-dynamic scenes.  Furthermore, for RoDUS to yield good reconstruction, it relies on a substantial number of input images from multiple views, making it less effective in few-shot settings.

\end{document}